\documentclass{article} 
\usepackage{vera_preprint,times}

\usepackage{amsmath,amsfonts,bm}

\def\eqref#1{equation~\ref{#1}}

\def\1{\bm{1}}

\DeclareMathAlphabet{\mathsfit}{\encodingdefault}{\sfdefault}{m}{sl}
\SetMathAlphabet{\mathsfit}{bold}{\encodingdefault}{\sfdefault}{bx}{n}

\ifdefined\XeTeXversion
  \providecommand{\DeclareUnicodeCharacter}[2]{}
\fi

\def\bm{\mathbf{m}}

\def\smskip{\smallskip}

\def\texitem#1{\par\smskip\noindent\hangindent 25pt
               \hbox to 25pt {\hss #1 ~}\ignorespaces}

\newcommand{\BEAS}{\begin{eqnarray*}}
\newcommand{\EEAS}{\end{eqnarray*}}
\newcommand{\BEA}{\begin{eqnarray}}
\newcommand{\EEA}{\end{eqnarray}}
\newcommand{\BEQ}{\begin{eqnarray}}
\newcommand{\EEQ}{\end{eqnarray}}
\newcommand{\BIT}{\begin{itemize}}
\newcommand{\EIT}{\end{itemize}}
\newcommand{\BNUM}{\begin{enumerate}}
\newcommand{\ENUM}{\end{enumerate}}

\newcommand{\BA}{\begin{array}}
\newcommand{\EA}{\end{array}}

\def\Pr{\mathbb{P}}

\usepackage{xcolor}

\newif\ifpagenumbering
\pagenumberingtrue

\pagenumberingfalse
\newsavebox{\theorembox}
\newsavebox{\lemmabox}
\newsavebox{\defnbox}
\newsavebox{\assbox}
\savebox{\theorembox}{\noindent\bf Theorem}
\savebox{\lemmabox}{\noindent\bf Lemma}
\savebox{\defnbox}{\noindent\bf Definition}

\usepackage{soul}
\usepackage[normalem]{ulem}
\DeclareUnicodeCharacter{2212}{~}

\usepackage{hyperref}
\usepackage{cleveref}
\crefname{assumption}{Assumption}{Assumptions}
\crefname{theorem}{Theorem}{Theorems}
\crefname{lemma}{Lemma}{Lemmas}

\usepackage{relsize}

\colorlet{yhclr}{orange!50!yellow}

\def \yh #1{\textcolor{yhclr}{#1}}
\def \yhrmv #1{\textcolor{yhclr}{\sout{#1}}}

\usepackage{booktabs}      
\usepackage{multirow}      
\usepackage{array}         
\usepackage{graphicx}      
\usepackage{amssymb}
\usepackage{amsthm}
\newtheorem{definition}{Definition}
\newtheorem{remark}{Remark}

\usepackage{algorithm}
\usepackage{algpseudocode} 
\usepackage{enumitem}
\usepackage{todonotes}
\definecolor{revisionblue}{RGB}{0,0,0}
\definecolor{revisionteal}{RGB}{0,0,0}
\definecolor{revisionviolet}{RGB}{0,0,0}
\definecolor{revisionorange}{RGB}{0,0,0}
\definecolor{proofreadmagenta}{RGB}{0,0,0}
\definecolor{reviewretention}{RGB}{0,0,0}
\definecolor{reviewsummary}{RGB}{0,0,0}
\definecolor{reviewfuture}{RGB}{0,0,0}
\definecolor{reviewconsistency}{RGB}{0,0,0}
\newif\ifarxivhighlight
\arxivhighlightfalse
\definecolor{arxivblue}{RGB}{0,70,160}
\newcommand{\arxivedit}[1]{\ifarxivhighlight\textcolor{arxivblue}{#1}\else#1\fi}
\renewcommand{\headrulewidth}{0.4pt}

\newcommand{\vera}{\texttt{VERA}}

\usepackage{hyperref}
\usepackage{url}

\title{VERA: Verdict-Conditioned Reliability for Adaptive LLM Judges}

\author{\arxivedit{Qiushui Xu$^{1}$}\thanks{\arxivedit{This work was done during an internship at Amazon.}}\hspace{0.4em}\arxivedit{,}
\arxivedit{Syamil Mohd Razak$^{2}$, Tao Yuan$^{2}$, Piotr Habas$^{2}$} \\
\arxivedit{$^{1}$ Penn State University, $^{2}$ Amazon} \\
\arxivedit{\texttt{qjx5019@psu.edu}},\ \arxivedit{\texttt{\{rsyamil,yuanty,phabas\}@amazon.com}}
}

\hypersetup{
  colorlinks=false,
  pdfborder={0 0 1},
  citebordercolor={0 1 0},
  pdftitle={VERA: Verdict-Conditioned Reliability for Adaptive LLM Judges},
  pdfauthor={Qiushui Xu, Syamil Mohd Razak, Tao Yuan, Piotr Habas},
  pdfsubject={Verdict-conditioned reliability for adaptive LLM judges},
  pdfkeywords={LLM judges, reliability estimation, continual adaptation}
}

\begin{document}

\maketitle

\begin{abstract}
Accurately estimating judgment reliability is a central challenge in adapting
LLM judges to newly verified feedback while preserving previously learned
behavior.
However, existing approaches often rely on output-level confidence, which can
be overconfident and poorly aligned with judgment correctness.
We propose \texttt{VERA}, a \emph{\textbf{VE}rdict-conditioned
\textbf{R}eliability \textbf{A}xis} that estimates reliability from hidden
activations by distinguishing correct from incorrect judgments within each
predicted-verdict group.
Using \texttt{VERA} as a control signal, we develop a \texttt{VERA}-guided periodic adaptation framework
that integrates reliability-ranked corrective updates, reliability-residual
replay, and periodic refresh of the reliability directions.
After \texttt{VERA}-guided adaptation on Chatbot Arena, 8B- and 14B-parameter
judges outperform the strongest baseline on each of four held-out
public benchmarks, with relative gains of up to \textbf{23.01\%}.
The framework also improves focal-class recall by up to
\textbf{16.1\%} relative to the strongest adaptive baselines
on a separate proprietary temporal auditing task.



\end{abstract}

\section{Introduction}
\label{sec:introduction}

Large language models (LLMs) are increasingly used as judges to rank candidate
responses~\citep{zheng2023judgingllmasajudgemtbenchchatbot}, provide reward
signals for model alignment~\citep{yuan2024selfrewarding}, curate training
data~\citep{majumdar2025geneticinstruct}, and automate
evaluation~\citep{liu2023geval}.
Supervised instruction fine-tuning has further enabled specialized open-weight
evaluators~\citep{zhu2025judgelmfinetunedlargelanguage,
kim2024prometheus2opensource}, while iterative preference learning can improve
both response generation and judging
capabilities~\citep{yuan2024selfrewarding}.
\emph{Adaptive judging} extends evaluation beyond a fixed checkpoint by
periodically updating a trained judge with newly available verified judgments.
This feedback provides opportunities to correct recurring errors and improve
performance on subsequently encountered cases.
However, repeated fine-tuning can degrade previously acquired behavior,
creating a need to incorporate new feedback while retaining useful prior
capabilities~\citep{wange2024inscl}.

\begin{figure*}[t]
    \centering
    \includegraphics[width=0.92\textwidth]{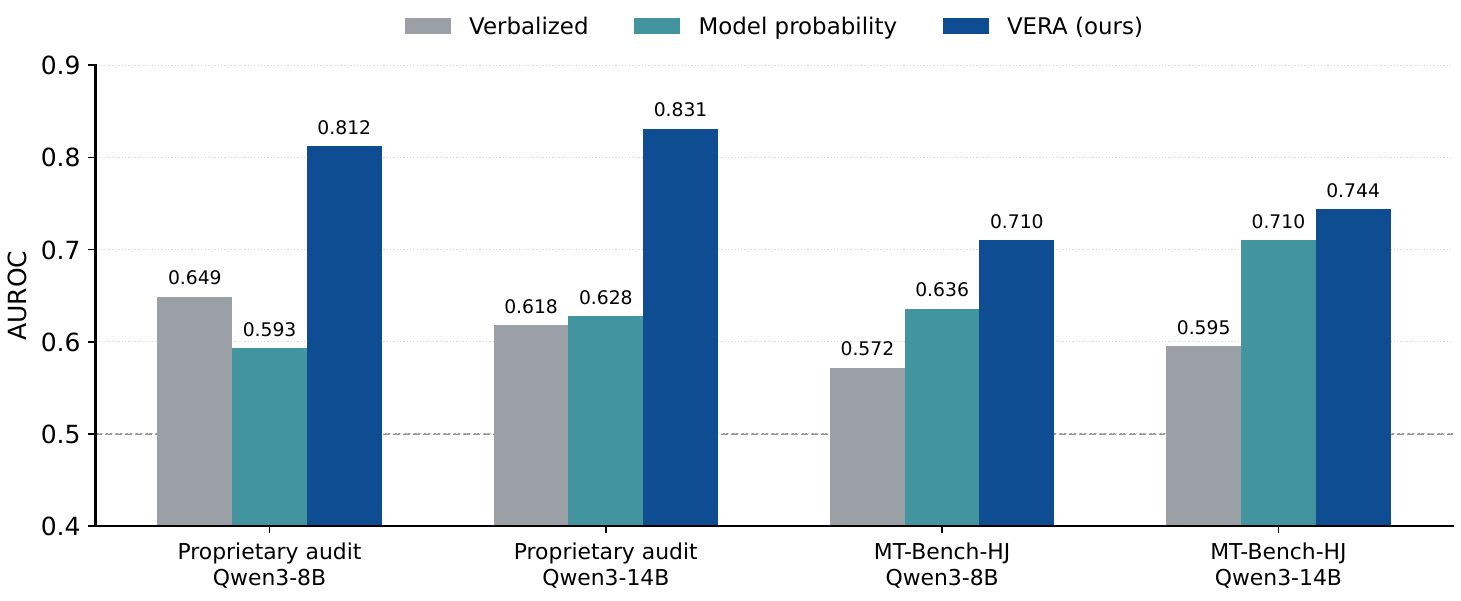}
    \caption{
    \textbf{Reliability signals at frozen checkpoints.}
    AUROC for distinguishing correct from incorrect judgments using verbalized
    confidence, model probability, and \vera, evaluated on the proprietary
    auditing task and MT-Bench-HJ with Qwen3-8B and Qwen3-14B.
    {\color{reviewconsistency}
    Values and evaluation conventions follow
    Table~\ref{tab:main-reliability}.}
    The dashed line denotes chance-level discrimination.
    }
    \label{fig:internal-reliability}
\end{figure*}

A central challenge in this setting is estimating \emph{judgment reliability}:
the probability that the judge's predicted verdict agrees with a verified
reference judgment~\citep{tian2023justask,
guo2017calibrationmodernneuralnetworks}.
An estimate of this quantity can inform which decisions warrant stronger
correction and which labeled examples should be retained for future updates.
Existing approaches obtain confidence estimates through verbalized
self-assessment~\citep{tian2023justask,xiong2024uncertainty},
model probabilities~\citep{tian2023justask}, repeated sampling and
consistency~\citep{xiong2024uncertainty,manakul2023selfcheckgpt},
or post-hoc calibration~\citep{guo2017calibrationmodernneuralnetworks}.
However, these estimates need not faithfully track verdict correctness:
a model can strongly prefer one verdict without that verdict agreeing with the
reference.
For the frozen judges in Figure~\ref{fig:internal-reliability}, verbalized
confidence provides weak correctness discrimination, while the informativeness
of model probabilities varies across tasks and model scales.
These observations motivate our central question:
\emph{can we obtain a more faithful estimate of judgment reliability and use
it to guide how judges learn from newly verified feedback?}

To address this question, we introduce \vera, a
\emph{\textbf{VE}rdict-conditioned \textbf{R}eliability \textbf{A}xis}
that estimates judgment reliability from hidden activations.
Rather than pooling all judgments into a single correctness direction,
\vera~compares correct and incorrect judgments within each predicted-verdict
group.
This conditioning accounts for verdict-specific representation structure and
reduces the risk of conflating differences between verdicts with differences
in their reliability.
For a new judgment, \vera~projects its hidden activation onto the direction
associated with the predicted verdict, producing a reliability score from the
same forward pass used for judging.
Figure~\ref{fig:internal-reliability} shows that this signal achieves higher
correctness-ranking AUROC than verbalized confidence and model probabilities
across the four evaluated task--model configurations.

Using this signal, we develop a \vera-guided periodic adaptation framework.
Within-verdict reliability ranks modulate corrective updates, while the
{\color{reviewconsistency}
discrepancy between post-update reliability estimates and observed correctness
defines a \emph{reliability residual} for prioritizing replay.}
We periodically refresh the reliability directions as the judge evolves.
Verified verdicts remain the supervision targets; \vera~controls how feedback
is weighted and retained.
This connects reliability estimation with adaptation, allowing us to assess
both signal quality and its utility for improving judges across diverse tasks. 

Our key contributions are summarized as follows:
\begin{itemize}

    \item \textbf{Verdict-conditioned internal reliability.}
    We introduce \vera, a lightweight, single-pass reliability signal that
    distinguishes correct from incorrect judgments within each predicted-verdict
    group.
    At frozen checkpoints, \vera~provides stronger correctness discrimination
    than verbalized confidence and model probabilities across two judging tasks
    and both Qwen3-8B and Qwen3-14B.
\vspace{-2mm}
    \item \textbf{\vera-guided periodic adaptation.}
    We develop an adaptation framework that uses internal reliability to
    modulate corrective updates, prioritize replay through reliability
    residuals, and periodically refresh the reliability directions.
    On a proprietary human-verified temporal auditing task, the framework
    improves focal-class recall by \textbf{16.1\%} and \textbf{8.8\%}
    relative to the strongest adaptive baseline at Qwen3-8B and Qwen3-14B,
    respectively.
\vspace{-2mm}
    \item \textbf{Cross-benchmark generalization and empirical analysis.}
    We evaluate judges initialized through JudgeLM instruction
    fine-tuning~\citep{zhu2025judgelmfinetunedlargelanguage} and periodically
    adapted on Chatbot Arena~\citep{zheng2023judgingllmasajudgemtbenchchatbot}.
    The adapted judges outperform the strongest evaluated baseline on each of
    four held-out public benchmarks:
    MT-Bench-HJ~\citep{zheng2023judgingllmasajudgemtbenchchatbot},
    Auto-J~\citep{li2024generativejudge},
    JudgeBench~\citep{tan2025judgebench}, and
    LLMBar~\citep{zeng2024llmbar}, with relative gains of up to
    \textbf{8.92\%} and \textbf{23.01\%} at Qwen3-8B and
    Qwen3-14B, respectively.
    Additional analyses on RewardBench~2~\citep{malik2026rewardbench2}
    characterize domain-specific Safety degradation and the benefits and
    costs of reliability-guided corrective replay.

\end{itemize}

\section{Related Work}
\label{sec:related}

\paragraph{Learned and adaptive LLM judges.}
LLM-as-a-judge methods provide scalable evaluation of open-ended model
responses~\citep{zheng2023judgingllmasajudgemtbenchchatbot}.
Specialized open-weight evaluators further improve judging capabilities
through supervised judgment data and task-specific evaluation
criteria~\citep{zhu2025judgelmfinetunedlargelanguage,
kim2024prometheus2opensource,leng2025praetor}.
Beyond supervised specialization, Self-Rewarding Language
Models~\citep{yuan2024selfrewarding} use model-generated judgments as rewards
for iterative preference optimization, improving both instruction following
and evaluation capabilities.
These approaches establish that judging capabilities can be learned and
iteratively refined.
Our focus is on how an existing judge incorporates newly verified feedback:
reference verdicts remain the supervision targets, while the judge's internal
reliability determines how that supervision is weighted and retained.


\paragraph{Confidence and internal reliability.}
LLM confidence estimation draws on verbalized
self-assessment~\citep{tian2023justask,xiong2024uncertainty},
model probabilities~\citep{tian2023justask}, sampling-based
consistency~\citep{xiong2024uncertainty,manakul2023selfcheckgpt}, and
post-hoc calibration~\citep{guo2017calibrationmodernneuralnetworks}.
Internal representations provide a complementary source of correctness
information.
Mahaut et al.~\citep{mahaut2024factualconfidence} find supervised
hidden-state probes effective for factual confidence estimation, while
InternalInspector~\citep{beigi2024internalinspector} uses contrastive learning
over internal states across transformer layers.
The concurrent Value Axis study~\citep{jiang2026valueaxis} constructs a
difference-in-means direction from synthetic in-context reinforcement-learning
trajectories to analyze and steer confidence and self-correction.
This activation-direction perspective motivates our representation, but
\vera~targets the correctness of an evaluator's discrete verdict rather than
the expected success of a task-solving trajectory.
It conditions the direction on the predicted verdict and uses the resulting
signal to guide corrective updates and replay, extending reliability estimation
from assessing judgments to controlling subsequent adaptation.

\paragraph{Sequential adaptation and replay.}
Sequential language-model adaptation seeks to incorporate new data while
preserving previously acquired capabilities.
Continual-T0~\citep{scialom2022continual} studies sequential acquisition of
language-generation tasks, while replay-based methods address which historical
examples should accompany subsequent updates.
InsCL~\citep{wange2024inscl} guides replay using instruction-based task
similarity and measures of instruction complexity and diversity.
KPIG~\citep{he2025kpig} uses key-part information gain to select replay
data and refine the instruction-tuning objective.
Our setting retains the judging task and verdict semantics while incorporating
periodically arriving verified feedback.
Within this setting, \vera~prioritizes replay using a reliability residual:
the discrepancy between estimated reliability and observed correctness.
This criterion retains judgments whose outcomes disagree with the model's
reliability estimate, rather than relying solely on instruction similarity or
error status.
Combined with reliability-ranked correction and periodic direction refresh,
it connects memory selection with the evolving judge's estimate of its own
correctness.

\vspace{-2mm}
\section{Methodology}
\label{sec:methodology}

We formalize judgment reliability in Section~\ref{sec:problem-formulation},
introduce \vera~in Section~\ref{sec:vera-directions}, and describe
reliability-guided adaptation in Section~\ref{sec:vera-adaptation}.
Figure~\ref{fig:method-overview} summarizes the framework and the complete
procedure appears in Appendix~\ref{app:vera-algorithm}.

\subsection{Problem Formulation}
\label{sec:problem-formulation}

An LLM judge $\pi_\theta$ receives an evaluation input $x$ and predicts a
verdict from a finite set $\mathcal{Y}$.
Given a verified reference $y^\star$, its prediction and realized correctness
are
\begin{equation}
    \hat y_\theta(x)=\arg\max_{y\in\mathcal{Y}}\pi_\theta(y\mid x),
    \qquad
    r_\theta(x)=\mathbf{1}\{\hat y_\theta(x)=y^\star\}.
    \label{eq:correctness}
\end{equation}
Following the correctness-probability interpretation of
confidence~\citep{kadavath2022language,tian2023justask,
guo2017calibrationmodernneuralnetworks}, we define the quantity to be estimated.

\begin{definition}[Judgment Reliability]
\label{def:judgment-reliability}
For an input $x$ and fixed judge $\theta$, judgment reliability is
\begin{equation}
    R_\theta(x)
    =\Pr\!\left(Y^\star=\hat y_\theta(x)\mid X=x\right)
    =\mathbb{E}\!\left[r_\theta(X)\mid X=x\right],
    \label{eq:reliability}
\end{equation}
where $(X,Y^\star)$ follows the reference input--label distribution.
\end{definition}

\vspace{-2mm}
Reliability concerns agreement with the reference, whereas
$\max_y\pi_\theta(y\mid x)$ is the model probability assigned to its
prediction; the two need not coincide.
We use this reliability estimate to guide \emph{periodic adaptation}:
starting from an initial judge $\theta_0$, each verified batch
$\mathcal{D}_t$ and previously retained examples $\mathcal{M}_{t-1}$ are
used to obtain $\theta_t$.
The objective is to incorporate new feedback while preserving useful prior
behavior.

\begin{figure*}[t]
    \centering
    \includegraphics[width=\textwidth]{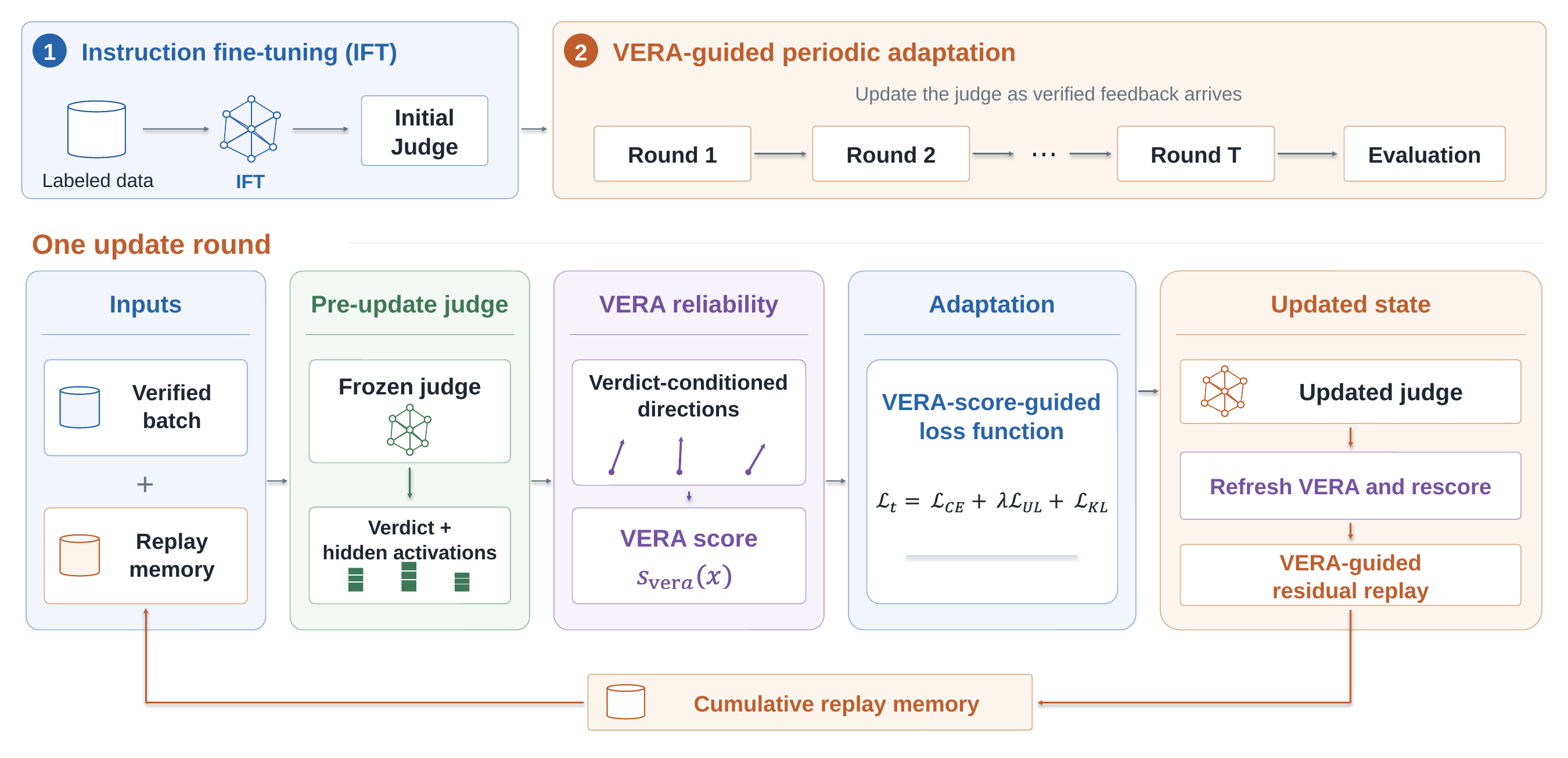}
    \caption{
    \textbf{Overview of \vera-guided periodic adaptation.}
    Pre-update reliability guides corrective training; the updated judge and
    current reliability estimator guide replay selection for later rounds.
    }
    \label{fig:method-overview}
\end{figure*}

\subsection{Verdict-Conditioned Reliability Directions}
\label{sec:vera-directions}

Let $h_\ell(x)\in\mathbb{R}^d$ denote the hidden activation at layer $\ell$
at the final prompt position, immediately before verdict generation.
A pooled comparison of correct and incorrect judgments can mix correctness
information with differences in predicted-verdict composition.
\vera~instead compares judgments within the same predicted-verdict group.

\begin{definition}[\vera~Direction]
\label{def:vera-direction}
For layer $\ell$ and predicted verdict $c\in\mathcal{Y}$, define
\begin{equation}
\begin{aligned}
    v_{\ell,c}
    ={}&\mathbb{E}[h_\ell(x)\mid\hat y_\theta(x)=c,r_\theta(x)=1]-\mathbb{E}[h_\ell(x)\mid\hat y_\theta(x)=c,r_\theta(x)=0].
\end{aligned}
    \label{eq:vera-axis}
\end{equation}
The expectations are estimated by sample means on verified training judgments.
\end{definition}

\begin{remark}[Global Reliability Direction]
\label{rem:global-vera}
Removing the predicted-verdict conditioning in
Definition~\ref{def:vera-direction} yields a global reliability direction:
\begin{equation}
    v_\ell^{\mathrm{global}}
    =
    \mathbb{E}[h_\ell(x)\mid r_\theta(x)=1]
    -
    \mathbb{E}[h_\ell(x)\mid r_\theta(x)=0].
    \label{eq:global-vera-axis}
\end{equation}
This construction pools judgments across all predicted verdicts.
When verdict composition differs between correct and incorrect judgments,
the resulting direction may capture verdict-specific structure in addition
to correctness.
We use this pooled construction as a comparison to assess the effect of
verdict conditioning. More details can be found in Table~\ref{tab:app-hidden-state-estimators}.
\end{remark}

The difference of means captures correctness-related variation while holding
the predicted verdict fixed, without training an auxiliary neural probe.
We convert this direction into a scalar score using cosine similarity.

\begin{definition}[\vera~Score]
\label{def:vera-score}
For input $x$, the \emph{\vera~score} is
\begin{equation}
    s_{\vera}(x)
    =\cos\!\left(h_{\ell^\star}(x),
        v_{\ell^\star,\hat y_\theta(x)}\right),
    \label{eq:vera-score}
\end{equation}
where $\ell^\star$ is the selected representation layer.
\end{definition}



Cosine similarity measures alignment with the corresponding reliability
direction rather than activation magnitude.
For each candidate layer $\ell\in\mathcal{L}$, we estimate the directions
on the training split and select the layer that maximizes correctness
separation on a disjoint validation split:
\begin{equation}
    \ell^\star
    =
    \arg\max_{\ell\in\mathcal{L}}
    \mathcal{J}_{\mathrm{val}}(\ell),
    \label{eq:vera-layer}
\end{equation}
where $\mathcal{J}_{\mathrm{val}}$ measures how well the layer's reliability
scores distinguish correct from incorrect judgments.
We use raw-score AUROC for proprietary judging and within-verdict mean-rank
separation for public judging.
The selected layer is shared across verdicts and fixed throughout periodic
adaptation.
The exact criteria and layer sweeps are reported in
Appendix~\ref{app:layer-selection-analysis}.

The raw score is not a correctness probability.
For adaptation, we use two transformations:
\begin{align}
    q(x)&=\widehat F_{\hat y_\theta(x)}\!\left(s_{\vera}(x)\right),
    \label{eq:vera-rank}\\
    V(x)&=C_{\hat y_\theta(x)}\!\left(s_{\vera}(x)\right).
    \label{eq:vera-value}
\end{align}
Here $\widehat F_c$ is the empirical CDF within predicted-verdict group $c$
of the current batch, and $C_c$ is a verdict-specific calibration map.
The rank $q(x)\in[0,1]$ expresses relative reliability for corrective
weighting, whereas $V(x)\in[0,1]$ estimates correctness probability for
residual replay.
Implementation details appear in Appendix~\ref{app:vera-calibration}.

\subsection{\vera-Guided Periodic Adaptation}
\label{sec:vera-adaptation}

At round $t$, we train on
$\mathcal{S}_t=\mathcal{D}_t\cup\mathcal{M}_{t-1}$.
Pre-update reliability guides correction, while post-update reliability
identifies examples to retain for subsequent rounds.

\paragraph{Reliability-ranked correction.}
Let $\hat y_t^-(x)=\hat y_{\theta_{t-1}}(x)$ and $q_t^-(x)$ denote the
pre-update verdict and within-verdict rank.
We assign larger corrective weights to lower-ranked judgments:
\begin{equation}
    \widetilde g_t(x)
    =\frac{1+\alpha(1-q_t^-(x))^2}
    {\mathbb{E}_{z\in\mathcal{D}_t}[1+\alpha(1-q_t^-(z))^2]},
    \qquad \alpha\geq0.
    \label{eq:vera-gate}
\end{equation}
The quadratic gate smoothly emphasizes the low-reliability tail, while
normalization keeps its batch-average weight at one.
We optimize
\begin{align}
    \mathcal{L}_t(\theta)
    &=\mathcal{L}_{\mathrm{CE}}
      +\lambda\mathcal{L}_{\mathrm{UL}}+\mathcal{L}_{\mathrm{KL}},
    \label{eq:vera-loss}\\
    \mathcal{L}_{\mathrm{UL}}
    &=\mathbb{E}_{x\in\mathcal{D}_t}\!\left[
        \widetilde g_t(x)\rho(x)
        \mathbf{1}\{\hat y_t^-(x)\neq y^\star\}\,
        u_t(x;\theta)\right],
    \label{eq:vera-ul}
\end{align}
where $u_t$ is an unlikelihood penalty on the previously incorrect verdict,
$\rho(x)\geq1$ is a task-specific error weight, and $\lambda\geq0$
controls the corrective term.
Weighted cross-entropy $\mathcal{L}_{\mathrm{CE}}$ learns from verified
labels in $\mathcal{S}_t$, while $\mathcal{L}_{\mathrm{KL}}$ discourages
unnecessary deviation from $\theta_{t-1}$.
Reference predictions, ranks, and weights remain fixed during optimization.
The full loss specification appears in Appendix~\ref{app:vera-objective}.

\paragraph{Periodic reliability refresh.}
Because adaptation changes the hidden representations, we periodically
re-estimate directions using the updated judge $\theta_t$ on
$\mathcal{D}_t$.
At the fixed layer $\ell^\star$, we smooth the new and previous directions:
\begin{equation}
    v_{c,t}=\operatorname{Normalize}\!\left(
        \gamma v^{\mathrm{new}}_{c,t}+(1-\gamma)v_{c,t-1}\right),
    \label{eq:vera-refresh}
\end{equation}
where both input directions are normalized before averaging and
$\gamma\in[0,1]$ controls the contribution of the new estimate.
Smoothing balances tracking the updated judge with retaining earlier
reliability information.
The calibration maps are also refitted on refresh rounds; otherwise,
directions and maps remain unchanged.

\paragraph{Reliability-residual replay.}
After training and any scheduled refresh, we evaluate the current batch under
$\theta_t$ and compute
\begin{equation}
    \Delta_{\vera,t}(x)=\left|r_{\theta_t}(x)-V_t(x)\right|,
    \label{eq:vera-residual}
\end{equation}
where $V_t(x)$ uses the updated judge and current reliability estimator.
The residual is large for an incorrect judgment estimated to be reliable or a
correct judgment estimated to be unreliable, prioritizing disagreement beyond
error status alone.
We add the $K_t$ highest-residual examples to the cumulative replay memory:
\begin{equation}
    \mathcal{C}_t=\operatorname{Top}_{K_t}
    (\mathcal{D}_t;\Delta_{\vera,t}),
    \qquad
    \mathcal{M}_t=\mathcal{M}_{t-1}\cup\mathcal{C}_t.
    \label{eq:vera-buffer}
\end{equation}
The union retains distinct examples, and $K_t$ is the per-round selection
budget.
The initial memory contains residual-selected historical examples available
before the first update.
Selected examples remain available to later rounds, preserving access to
judgments that previously disagreed with the reliability estimate.

\vspace{-2mm}
\section{Experiments}
\label{sec:experiments}

We evaluate \vera~on public judging benchmarks and a proprietary temporal
auditing task, examining both adaptation performance and judgment-reliability
estimation.
We first describe the experimental setup, then present results for each judging
setting and compare reliability signals across training stages.

\subsection{Experimental Setup}
\label{sec:experimental-setup}

\subsubsection{Datasets}
\label{sec:exp-datasets}

\paragraph{Public judging benchmarks.}
We initialize judges through instruction fine-tuning on
JudgeLM~\citep{zhu2025judgelmfinetunedlargelanguage} and construct five
periodic feedback batches from Chatbot
Arena~\citep{zheng2023judgingllmasajudgemtbenchchatbot}.
We evaluate the resulting judges on four held-out benchmarks:
MT-Bench-HJ~\citep{zheng2023judgingllmasajudgemtbenchchatbot},
Auto-J Eval-P~\citep{li2024generativejudge},
JudgeBench~\citep{tan2025judgebench}, and
LLMBar~\citep{zeng2024llmbar}.
MT-Bench-HJ contains $6{,}710$ ordered examples, and Auto-J contains
$1{,}392$ response pairs; both include A-win, B-win, and tie verdicts.
We use the $350$-item GPT-4o subset of JudgeBench for objective-correctness
evaluation and the four adversarial subsets of LLMBar for instruction-following
evaluation.
RewardBench~2~\citep{malik2026rewardbench2} provides an additional
cross-format evaluation and Safety-retention analysis.
{\color{revisionblue}
The JudgeLM training set contains $99{,}647$ response pairs, and the five-batch Arena feedback
stream contains $31{,}340$ comparisons.
}

\paragraph{Proprietary temporal auditing.}
We additionally evaluate a binary auditing task with approximately $65$K
human-verified historical examples followed by periodic feedback.
An earlier holdout measures \emph{anchor retention}, while a later window
measures \emph{forward performance} under a substantial change in class
prevalence.
For confidentiality, we denote the verdicts by \textsc{Critical} and
\textsc{Non-Critical}.
The number of feedback batches, anchor-set size, and forward-set size are
{\color{revisionblue}4, $2{,}634$, and $894$, respectively}.

\vspace{-2mm}
\subsubsection{Metrics}
\label{sec:exp-metrics}

We distinguish verdict quality from reliability discrimination.
Verdict-quality metrics include accuracy, macro-F1, and focal-class recall
and F1.
The focal class is \textsc{Critical} in the proprietary task and the tie
verdict in three-way preference judging.
For public benchmarks, we retain the respective evaluation metrics:
row accuracy for MT-Bench-HJ, two-order agreement and consistency for
Auto-J, two-order accuracy for JudgeBench, and macro-averaged accuracy across
the four adversarial subsets of LLMBar. More details can be found in Appendix~\ref{app:mixed-format}.
For reliability estimation, we report AUROC for distinguishing correct from
incorrect judgments and explicitly identify the confidence signal being evaluated.
Anchor performance and source-validation accuracy are used to assess retention.
Relative gains over a baseline score $b$ are computed as
$100(s-b)/b\%$, where $s$ is the corresponding \vera~score.
Absolute changes are explicitly identified.
{\color{revisionblue}
In public three-way judging, \texttt{C} denotes a tie and is scored as a
regular verdict.
In proprietary auditing, \texttt{C} denotes abstention and is retained in the
evaluation denominator as an incorrect judgment.
The candidate-token argmax used in the main experiments always returns one of
the valid verdicts, so invalid outputs do not occur.
}

\vspace{-2mm}
\subsubsection{Baselines}
\label{sec:exp-baselines}

For public judging, we compare against \textbf{IFT}, a judge
instruction-tuned on JudgeLM~\citep{zhu2025judgelmfinetunedlargelanguage},
and \textbf{DPO}, which further trains the IFT judge on Chatbot Arena
preferences~\citep{zheng2023judgingllmasajudgemtbenchchatbot,
rafailov2023directpreferenceoptimization}.
For proprietary auditing, we compare \textbf{periodic IFT},
\textbf{periodic DPO}, and \textbf{periodic
RAG}~\citep{lewis2020retrieval}, together with a \textbf{frozen initial judge}
that is not updated during the feedback stream.
Throughout the comparisons below, the strongest baseline is determined
separately for each metric and model scale: the better of IFT and DPO for
public judging, and the best of periodic IFT, DPO, and RAG for proprietary
adaptation.

\vspace{-2mm}
\subsubsection{Implementation Details}
\label{sec:exp-implementation}

We use Qwen3-8B and Qwen3-14B~\citep{yang2025qwen3} with unified
single-token verdicts matched to each task's label space.
Reliability directions are constructed from training examples, and
representation layers are selected using disjoint validation data.
{\color{revisionblue}
The proprietary task uses validation correctness AUROC, whereas public
adaptation uses correct--incorrect separation after within-verdict rank
normalization, matching the score used by its corrective objective.
}
{\color{revisionblue}
The forward set is a strictly independent test set used only for final
evaluation.}
{\color{revisionblue}
For periodic adaptation, we fix the IFT-selected zero-indexed layers at $20$
and $38$ for the proprietary 8B and 14B judges, and at $33$ and $32$ for their
public counterparts.
}
Each selected layer remains fixed across subsequent checkpoints and reliability
refreshes.
The proprietary \vera~configurations use replay fractions $f=0.010$ at
8B and $f=0.015$ at 14B.
{\color{revisionblue}
The public replay buffer initially contains $740$ and $1{,}480$ selected JudgeLM
items and adds $240$ and $480$ items from each observed Arena batch at 8B and 14B respectively; both response orders are retained for every selected item, and
the replay weight is $0.15$.
Initial IFT uses a learning rate of $10^{-4}$, an effective batch size of
$256$, and two epochs; DPO uses $5\times10^{-6}$, $128$, and one epoch;
each \vera-guided periodic update uses $5\times10^{-5}$, $128$, and one
epoch per feedback batch.
The proprietary RAG baseline keeps the IFT judge fixed and retrieves the top
three examples from a FAISS cosine-similarity index over labeled feedback
batches observed up to the current period, using sentence-transformer
embeddings and excluding the query itself by identifier.
All training and adaptation hyperparameters are fixed before evaluation on the
forward set. More details can be found in Appendix~\ref{app:hyperparameters}.
}

\vspace{-2mm}
\subsection{Public Judging Benchmarks}
\label{sec:public-results}
\label{sec:main-results}

We first examine whether \vera-guided adaptation improves performance across
different judging tasks.
Table~\ref{tab:main-cross-dataset} compares IFT, DPO, and \vera~at both
model scales.
Within each method, the same resulting checkpoint is evaluated across all four
benchmarks.

\begin{table}[t]
    \centering
    \small
    \setlength{\tabcolsep}{6pt}
    \begin{tabular}{llrrrr}
        \toprule
        Model & Method & MT-Bench-HJ (\%) & Auto-J(\%) & JudgeBench(\%) & LLMBar(\%) \\
        \midrule
        \multirow{3}{*}{Qwen3-8B}
        & IFT
        & $64.42 \pm 0.13$ & $58.93 \pm 0.61$ & $56.48 \pm 0.59$ & $56.06 \pm 0.37$ \\
        & DPO
        & $64.57 \pm 0.19$ & $58.93 \pm 0.49$ & $57.06 \pm 0.32$ & $59.63 \pm 0.12$ \\
        & \textbf{\vera}
        & \textbf{66.69 $\pm$ 0.36} & \textbf{59.85 $\pm$ 1.01}
        & \textbf{57.22 $\pm$ 1.86} & \textbf{64.95 $\pm$ 2.40} \\
        \multicolumn{2}{l}{Relative gain over best baseline }
        & $+3.28$ & $+1.56$ & $+0.28$ & $+8.92$ \\
        \midrule
        \multirow{3}{*}{Qwen3-14B}
        & IFT
        & $64.10 \pm 0.21$ & $60.69 \pm 0.17$ & $60.20 \pm 1.17$ & $54.95 \pm 0.58$ \\
        & DPO
        & $64.50 \pm 0.13$ & $60.85 \pm 0.25$ & $60.46 \pm 1.31$ & $57.05 \pm 1.25$ \\
        & \textbf{\vera}
        & \textbf{67.73 $\pm$ 0.24} & \textbf{62.54 $\pm$ 0.43}
        & \textbf{62.08 $\pm$ 1.02} & \textbf{70.18 $\pm$ 1.12} \\
        \multicolumn{2}{l}{Relative gain over best baseline}
        & $+5.01$ & $+2.78$ & $+2.68$ & $+23.01$ \\
        \bottomrule
    \end{tabular}
    \caption{
    \textbf{Performance on four held-out public judging benchmarks (\%).}
    MT-Bench-HJ reports row accuracy; Auto-J reports two-order agreement;
    JudgeBench reports two-order accuracy on its $350$-item GPT-4o subset;
    LLMBar reports the macro-average over four adversarial subsets.
    Entries report means and sample standard deviations over three seeds.
    Gains are relative improvements (\%) over the stronger of IFT
    and DPO within each model--benchmark comparison.
    Bold marks the best score within each model scale.
    Scores should be compared within columns rather than averaged across
    benchmarks with different evaluation metrics.
    }
    \label{tab:main-cross-dataset}
\end{table}

{\color{reviewretention}
\vera~achieves the highest observed mean score on all four benchmarks at both scales.
}
At Qwen3-8B, relative improvements over the stronger baseline range from
$0.28\%$ to $8.92\%$; at Qwen3-14B, they range from $2.68\%$ to $23.01\%$.
The largest gain occurs on adversarial instruction following: \vera~raises
LLMBar accuracy from $59.63\%$ to $64.95\%$ at 8B and from $57.05\%$ to
$70.18\%$ at 14B.

The gains extend beyond pairwise preference judging to objective correctness
and adversarial instruction following, supporting cross-benchmark
generalization of the adapted judge.
{\color{reviewretention}
Their magnitudes remain task-dependent: on 8B JudgeBench, the means are
$57.22\%$ for \vera~and $57.06\%$ for DPO, while the \vera~seed standard
deviation is $1.86\%$. This is an observed mean difference, not a claim of
statistical significance; the adversarial gains are substantially larger.
}
The 14B judge shows larger relative gains than 8B on each benchmark, although
these comparisons alone do not identify which component of the framework
produces the improvement.

\vspace{-2mm}
\subsection{Proprietary Temporal Auditing}
\label{sec:proprietary-results}

We next evaluate adaptation on the proprietary feedback stream.
Table~\ref{tab:main-proprietary-adaptation} reports critical-class recall and
F1 for the final judge on the earlier anchor and later forward sets,
allowing adaptation and retention to be examined together.

\begin{table}[t]
    \centering
    \small
    \setlength{\tabcolsep}{4pt}
    \begin{tabular}{llrrrr}
        \toprule
        Model & Method
        & \multicolumn{2}{c}{Earlier anchor}
        & \multicolumn{2}{c}{Forward set} \\
        \cmidrule(lr){3-4}\cmidrule(lr){5-6}
        & & Recall (\%) & F1 (\%) & Recall (\%) & F1 (\%) \\
        \midrule
        \multirow{5}{*}{Qwen3-8B}
        & Frozen initial judge
        & 85.8 & 78.8 & 65.7 & 58.8 \\
        & Periodic RAG
        & 82.0 & 79.0 & 61.6 & 61.6 \\
        & Periodic IFT
        & 72.5 & 77.4 & 57.6 & 64.0 \\
        & Periodic DPO
        & 84.1 & 78.7 & 62.6 & 57.9 \\
        & \textbf{\vera}
        & \textbf{86.0} & \textbf{79.1}
        & \textbf{72.7} & \textbf{66.1} \\
        \multicolumn{2}{l}{Relative gain over best baseline}
        & $+2.3$ & $+0.1$ & $+16.1$ & $+3.3$ \\
        \midrule
        \multirow{5}{*}{Qwen3-14B}
        & Frozen initial judge
        & 82.6 & 78.8 & 70.7 & 56.7 \\
        & Periodic RAG
        & 77.9 & 77.5 & 68.7 & \textbf{70.8} \\
        & Periodic IFT
        & 76.6 & 79.0 & 65.7 & 69.1 \\
        & Periodic DPO
        & 81.5 & 78.7 & 69.7 & 60.8 \\
        & \textbf{\vera}
        & \textbf{85.3} & \textbf{79.1} & \textbf{75.8} & 69.1 \\
        \multicolumn{2}{l}{Relative gain over best baseline}
        & $+4.7$ & $+0.1$ & $+8.8$ & $-2.4$ \\
        \bottomrule
    \end{tabular}
    \caption{
    \textbf{Periodic adaptation on the proprietary auditing task (\%).}
    Recall and F1 refer to the \textsc{Critical} verdict at the final update.
    \vera~uses replay fractions $f=0.010$ and $f=0.015$ at 8B and 14B,
    respectively.
    The frozen initial judge is task-trained but receives no subsequent
    feedback updates.
    Gains are relative improvements (\%) over the strongest
    adaptive baseline among periodic IFT, DPO, and RAG, determined separately
    for each model scale and metric.
    Bold marks the best reported score within each model scale and column.
    }
    \label{tab:main-proprietary-adaptation}
\end{table}

\paragraph{Forward performance.}
At Qwen3-8B, \vera~improves forward critical-class recall from $0.626$
under the strongest adaptive baseline, periodic DPO, to $0.727$, a relative
gain of \textbf{16.1\%}.
It also achieves the highest forward F1 ($0.661$ versus $0.640$).
At Qwen3-14B, periodic DPO provides the strongest baseline recall ($0.697$),
while \vera~reaches $0.758$, a relative improvement of \textbf{8.8\%}.
Thus, \vera~improves forward critical-class recall at both scales.

\paragraph{Retention and recall--F1 trade-offs.}
\vera~also achieves the highest anchor recall at both scales, reaching
$0.860$ at 8B and $0.853$ at 14B, compared with $0.858$ and $0.826$ for
the corresponding frozen initial judges.
Anchor F1 is $0.791$ at both scales, close to the frozen reference of $0.788$.
Periodic RAG has higher forward F1 at 14B ($0.708$ versus $0.691$),
but lower forward recall
($0.687$ versus $0.758$).
These results show improved critical-case coverage while maintaining the
reported anchor performance, with a forward recall--F1 trade-off at 14B.

\vspace{-2mm}
\subsection{Reliability Analysis}
\label{sec:reliability-analysis}

We finally compare the reliability signals used to distinguish correct from
incorrect judgments.
Table~\ref{tab:main-reliability} evaluates verbalized confidence, model
probabilities, and \vera~at frozen, IFT, and DPO checkpoints.
Here, \emph{Frozen} denotes the pre-IFT checkpoint, unlike the task-trained
frozen baseline in Table~\ref{tab:main-proprietary-adaptation}.

\begin{table}[t]
    \centering
    \small
    \setlength{\tabcolsep}{6pt}
    \begin{tabular}{lllrrr}
        \toprule
        Setting & Model & Stage & Verbalized & Softmax & \vera \\
        \midrule
        \multirow{6}{*}{Proprietary}
        & \multirow{3}{*}{8B}
        & Frozen & 0.649 & 0.593 & \textbf{0.812} \\
        & & IFT & 0.497 & 0.857 & \textbf{0.867} \\
        & & DPO & 0.513 & 0.852 & \textbf{0.866} \\
        \cmidrule(lr){2-6}
        & \multirow{3}{*}{14B}
        & Frozen & 0.618 & 0.628 & \textbf{0.831} \\
        & & IFT & 0.747 & 0.866 & \textbf{0.869} \\
        & & DPO & 0.779 & 0.870 & \textbf{0.872} \\
        \midrule
        \multirow{6}{*}{MT-Bench-HJ}
        & \multirow{3}{*}{8B}
        & Frozen & 0.572 & 0.636 & \textbf{0.710} \\
        & & IFT & 0.532 & \textbf{0.732} & 0.726 \\
        & & DPO & 0.528 & \textbf{0.736} & 0.727 \\
        \cmidrule(lr){2-6}
        & \multirow{3}{*}{14B}
        & Frozen & 0.595 & 0.710 & \textbf{0.744} \\
        & & IFT & 0.532 & 0.735 & \textbf{0.736} \\
        & & DPO & 0.537 & 0.730 & \textbf{0.735} \\
        \bottomrule
    \end{tabular}
    \caption{
    \textbf{AUROC for distinguishing correct from incorrect judgments.}
    Frozen denotes the pre-IFT checkpoint.
    MT-Bench-HJ verbalized confidence uses a separate generation pass,
    with correctness determined by that pass's verdict.
    Bold marks the largest AUROC or mean in each row.
    }
    \label{tab:main-reliability}
\end{table}

\paragraph{Reliability before task-specific training.}
On the frozen proprietary judges, \vera~exceeds softmax AUROC by $0.219$
at 8B and $0.203$ at 14B.
{\color{reviewretention}
The corresponding improvements on MT-Bench-HJ are $0.074$ and $0.034$.
}
\vera~also outperforms verbalized confidence in all four frozen
configurations.
The proprietary 14B result illustrates why reliability discrimination must
be distinguished from verdict quality: the judge exhibits near-single-verdict
predictions despite a strong AUROC.

\paragraph{Reliability after task-specific training.}
Following IFT and DPO, the difference between \vera~and softmax narrows to
an absolute gap of at most $0.014$ across the reported settings.
Verbalized confidence remains near chance on MT-Bench-HJ but becomes more
informative for the proprietary 14B judge.
{\color{proofreadmagenta}
These results identify cold-start judging as a particularly effective use
of \vera: verified judgments establish a reliability estimator before
optimizing the judge for the task.
After IFT or DPO, the estimator supplies the corrective weights and replay
scores used for further adaptation
(Appendices~\ref{app:internal-components} and~\ref{app:rb2-safety-retrieval}).
The signal also transfers to unseen judging tasks, and incorporating verified
target judgments further improves its discrimination
(Appendix~\ref{app:vera-transfer}).
}

\paragraph{Additional analyses.}
{\color{reviewsummary}
Appendix~\ref{app:additional-reliability} compares global directions,
verdict-conditioned \vera, and logistic-regression probes, and examines
layer selection and task transfer; verdict conditioning improves AUROC, while target-specific directions improve
average AUROC across public benchmarks.
{\color{reviewfuture}
For Qwen3-8B on the proprietary task, verdict conditioning raises AUROC
from $0.671$ to $0.812$ at the frozen checkpoint and from $0.848$ to
$0.867$ after IFT.}
Appendix~\ref{app:additional-adaptation} studies replay selection, budgets,
component ablations, and Safety repair; \vera~replay improves forward
recall over random and softmax-residual selection in the proprietary experiment and recovers more Safety accuracy than random replay at both scales.
Appendix~\ref{app:inference-scaffolds} evaluates training-free inference
harnesses; these changes do not consistently improve both accuracy and
verbalized-confidence AUROC on MT-Bench-HJ, while multi-call panels increase
latency over a single judge call in the proprietary evaluation. These findings further motivate using internal reliability signals to guide
adaptation beyond prompt and harness changes. {\color{reviewfuture}
Appendix~\ref{app:mixed-format} examines cross-format generalization and
benchmark placement across preference agreement, multi-way selection,
objective correctness, and adversarial instruction following.
\vera~achieves higher mean scores than IFT and DPO on all four pairwise
benchmarks at both scales, alongside Safety losses on RewardBench~2.
It is also competitive with published judge results: on JudgeBench,
14B \vera~achieves $62.08\%$ accuracy, compared with the published
$56.57\%$ for GPT-4o with Arena-Hard prompting.
}
}

\vspace{-2mm}
\section{Conclusion \& Future Work}
\label{sec:conclusion}

{\color{reviewfuture}
We introduced \vera, a verdict-conditioned internal reliability representation
for adaptive LLM judges.
\vera~uses this internal signal to rank corrective updates and replay
examples.
It improves critical-verdict recall on the proprietary task.
On public data, the same periodically adapted checkpoints improve over JudgeLM
IFT and Arena DPO on four public judge benchmarks at both model scales.
On RewardBench~2, a \vera-selected replay update recovers more held-out
Safety accuracy than random replay, with a small decrease in MT-Bench-HJ
accuracy.
Future work should investigate whether this reliability signal can guide
online reinforcement learning through feedback selection and update weighting.
Another direction is to track domain-level accuracy and reliability through
the update sequence, identify early signs of capability degradation, and
trigger targeted replay before further performance declines.
{\color{black}
Adapting feedback and replay budgets to these domain-level diagnostics could
help balance learning from new judgments with retaining earlier capabilities
under limited annotation budgets.}
}

\bibliography{iclr2027_conference}

@inproceedings{zheng2023judgingllmasajudgemtbenchchatbot,
  title     = {Judging {LLM}-as-a-Judge with {MT}-Bench and Chatbot Arena},
  author    = {Zheng, Lianmin and
               Chiang, Wei-Lin and
               Sheng, Ying and
               Zhuang, Siyuan and
               Wu, Zhanghao and
               Zhuang, Yonghao and
               Lin, Zi and
               Li, Zhuohan and
               Li, Dacheng and
               Xing, Eric P. and
               Zhang, Hao and
               Gonzalez, Joseph E. and
               Stoica, Ion},
  booktitle = {Advances in Neural Information Processing Systems},
  volume    = {36},
  year      = {2023},
  pages     = {46595--46623},
  publisher = {Curran Associates, Inc.},
  url       = {https://proceedings.neurips.cc/paper_files/paper/2023/hash/91f18a1287b398d378ef22505bf41832-Abstract-Datasets_and_Benchmarks.html}
}

@inproceedings{zhu2025judgelmfinetunedlargelanguage,
  title     = {{JudgeLM}: Fine-tuned Large Language Models are Scalable Judges},
  author    = {Zhu, Lianghui and
               Wang, Xinggang and
               Wang, Xinlong},
  booktitle = {The Thirteenth International Conference on Learning Representations},
  year      = {2025},
  url       = {https://proceedings.iclr.cc/paper_files/paper/2025/hash/7f8f73134e253845a8f82983219a8452-Abstract-Conference.html}
}

@inproceedings{kim2024prometheus2opensource,
  title     = {Prometheus 2: An Open Source Language Model Specialized in Evaluating Other Language Models},
  author    = {Kim, Seungone and
               Suk, Juyoung and
               Longpre, Shayne and
               Lin, Bill Yuchen and
               Shin, Jamin and
               Welleck, Sean and
               Neubig, Graham and
               Lee, Moontae and
               Lee, Kyungjae and
               Seo, Minjoon},
  booktitle = {Proceedings of the 2024 Conference on Empirical Methods in Natural Language Processing},
  month     = nov,
  year      = {2024},
  address   = {Miami, Florida, USA},
  publisher = {Association for Computational Linguistics},
  pages     = {4334--4353},
  doi       = {10.18653/v1/2024.emnlp-main.248},
  url       = {https://aclanthology.org/2024.emnlp-main.248/}
}

@inproceedings{malik2026rewardbench2,
  title     = {{RewardBench 2}: Advancing Reward Model Evaluation},
  author    = {Malik, Saumya and
               Pyatkin, Valentina and
               Land, Sander and
               Morrison, Jacob and
               Smith, Noah A. and
               Hajishirzi, Hannaneh and
               Lambert, Nathan},
  booktitle = {The Fourteenth International Conference on Learning Representations},
  year      = {2026}
}

@inproceedings{rafailov2023directpreferenceoptimization,
  title     = {Direct Preference Optimization: Your Language Model is Secretly a Reward Model},
  author    = {Rafailov, Rafael and
               Sharma, Archit and
               Mitchell, Eric and
               Manning, Christopher D. and
               Ermon, Stefano and
               Finn, Chelsea},
  booktitle = {Advances in Neural Information Processing Systems},
  volume    = {36},
  year      = {2023},
  publisher = {Curran Associates, Inc.},
  url       = {https://proceedings.neurips.cc/paper_files/paper/2023/hash/a85b405ed65c6477a4fe8302b5e06ce7-Abstract-Conference.html}
}

@inproceedings{guo2017calibrationmodernneuralnetworks,
  title     = {On Calibration of Modern Neural Networks},
  author    = {Guo, Chuan and
               Pleiss, Geoff and
               Sun, Yu and
               Weinberger, Kilian Q.},
  booktitle = {Proceedings of the 34th International Conference on Machine Learning},
  series    = {Proceedings of Machine Learning Research},
  volume    = {70},
  pages     = {1321--1330},
  year      = {2017},
  publisher = {PMLR},
  url       = {https://proceedings.mlr.press/v70/guo17a.html}
}

@article{yang2025qwen3,
  title   = {{Qwen3} Technical Report},
  author  = {Yang, An and others},
  journal = {arXiv preprint arXiv:2505.09388},
  year    = {2025}
}

@article{jiang2026valueaxis,
  title   = {The Value Axis: Language Models Encode Whether They're on the Right Track},
  author  = {Jiang, Nick and Kauvar, Isaac and Lindsey, Jack},
  journal = {arXiv preprint arXiv:2606.17056},
  year    = {2026},
  url     = {https://arxiv.org/abs/2606.17056}
}

@inproceedings{yuan2024selfrewarding,
  title     = {Self-Rewarding Language Models},
  author    = {Yuan, Weizhe and
               Pang, Richard Yuanzhe and
               Cho, Kyunghyun and
               Li, Xian and
               Sukhbaatar, Sainbayar and
               Xu, Jing and
               Weston, Jason E.},
  booktitle = {Proceedings of the 41st International Conference on Machine Learning},
  series    = {Proceedings of Machine Learning Research},
  volume    = {235},
  pages     = {57905--57923},
  year      = {2024},
  publisher = {PMLR}
}

@inproceedings{majumdar2025geneticinstruct,
  title     = {Genetic Instruct: Scaling up Synthetic Generation of Coding
               Instructions for Large Language Models},
  author    = {Majumdar, Somshubra and
               Noroozi, Vahid and
               Samadi, Mehrzad and
               Narenthiran, Sean and
               Ficek, Aleksander and
               Ahmad, Wasi Uddin and
               Huang, Jocelyn and
               Balam, Jagadeesh and
               Ginsburg, Boris},
  booktitle = {Proceedings of the 63rd Annual Meeting of the Association for
               Computational Linguistics (Volume 6: Industry Track)},
  pages     = {208--221},
  year      = {2025},
  publisher = {Association for Computational Linguistics}
}

@inproceedings{liu2023geval,
  title     = {{G}-Eval: {NLG} Evaluation using {GPT}-4 with Better Human Alignment},
  author    = {Liu, Yang and
               Iter, Dan and
               Xu, Yichong and
               Wang, Shuohang and
               Xu, Ruochen and
               Zhu, Chenguang},
  booktitle = {Proceedings of the 2023 Conference on Empirical Methods in
               Natural Language Processing},
  pages     = {2511--2522},
  year      = {2023},
  publisher = {Association for Computational Linguistics},
  doi       = {10.18653/v1/2023.emnlp-main.153}
}

@inproceedings{leng2025praetor,
  title     = {Praetor: A Fine-Grained Generative LLM Evaluator with
               Interpretable Output},
  author    = {Leng, Yichong and others},
  booktitle = {Proceedings of the 63rd Annual Meeting of the Association for
               Computational Linguistics (Volume 1: Long Papers)},
  year      = {2025},
  publisher = {Association for Computational Linguistics}
}

@inproceedings{wange2024inscl,
  title     = {{InsCL}: A Data-efficient Continual Learning Paradigm for
               Fine-tuning Large Language Models with Instructions},
  author    = {Wang, Yifan and
               Liu, Yafei and
               Shi, Chufan and
               Li, Haoling and
               Chen, Chen and
               Lu, Haonan and
               Yang, Yujiu},
  booktitle = {Proceedings of the 2024 Conference of the North American Chapter
               of the Association for Computational Linguistics:
               Human Language Technologies (Volume 1: Long Papers)},
  pages     = {663--677},
  year      = {2024},
  publisher = {Association for Computational Linguistics},
  doi       = {10.18653/v1/2024.naacl-long.37}
}

@inproceedings{tian2023justask,
    title = {Just Ask for Calibration: Strategies for Eliciting Calibrated
             Confidence Scores from Language Models Fine-Tuned with Human Feedback},
    author = {Tian, Katherine and
              Mitchell, Eric and
              Zhou, Allan and
              Sharma, Archit and
              Rafailov, Rafael and
              Yao, Huaxiu and
              Finn, Chelsea and
              Manning, Christopher},
    booktitle = {Proceedings of the 2023 Conference on Empirical Methods in
                 Natural Language Processing},
    month = dec,
    year = {2023},
    address = {Singapore},
    publisher = {Association for Computational Linguistics},
    pages = {5433--5442},
    doi = {10.18653/v1/2023.emnlp-main.330},
    url = {https://aclanthology.org/2023.emnlp-main.330/}
}

@inproceedings{xiong2024uncertainty,
    title = {Can LLMs Express Their Uncertainty? An Empirical Evaluation of
             Confidence Elicitation in LLMs},
    author = {Xiong, Miao and
              Hu, Zhiyuan and
              Lu, Xinyang and
              Li, Yifei and
              Fu, Jie and
              He, Junxian and
              Hooi, Bryan},
    booktitle = {The Twelfth International Conference on Learning Representations},
    year = {2024},
    url = {https://openreview.net/forum?id=gjeQKFxFpZ}
}

@inproceedings{manakul2023selfcheckgpt,
    title = {{SelfCheckGPT}: Zero-Resource Black-Box Hallucination Detection
             for Generative Large Language Models},
    author = {Manakul, Potsawee and
              Liusie, Adian and
              Gales, Mark},
    booktitle = {Proceedings of the 2023 Conference on Empirical Methods in
                 Natural Language Processing},
    month = dec,
    year = {2023},
    address = {Singapore},
    publisher = {Association for Computational Linguistics},
    pages = {9004--9017},
    doi = {10.18653/v1/2023.emnlp-main.557},
    url = {https://aclanthology.org/2023.emnlp-main.557/}
}

@inproceedings{li2024generativejudge,
  title     = {Generative Judge for Evaluating Alignment},
  author    = {Li, Junlong and
               Sun, Shichao and
               Yuan, Weizhe and
               Fan, Run-Ze and
               Zhao, Hai and
               Liu, Pengfei},
  booktitle = {The Twelfth International Conference on Learning Representations},
  year      = {2024},
  url       = {https://openreview.net/forum?id=gtkFw6sZGS}
}

@inproceedings{zeng2024llmbar,
  title     = {Evaluating Large Language Models at Evaluating Instruction Following},
  author    = {Zeng, Zhiyuan and
               Yu, Jiatong and
               Gao, Tianyu and
               Meng, Yu and
               Goyal, Tanya and
               Chen, Danqi},
  booktitle = {The Twelfth International Conference on Learning Representations},
  year      = {2024}
}

@inproceedings{tan2025judgebench,
  title     = {JudgeBench: A Benchmark for Evaluating LLM-Based Judges},
  author    = {Tan, Sijun and
               Zhuang, Siyuan and
               Montgomery, Kyle and
               Tang, William Y. and
               Cuadron, Alejandro and
               Wang, Chenguang and
               Popa, Raluca Ada and
               Stoica, Ion},
  booktitle = {The Thirteenth International Conference on Learning Representations},
  year      = {2025}
}

@inproceedings{mahaut2024factualconfidence,
  title     = {Factual Confidence of {LLM}s: on Reliability and Robustness of
               Current Estimators},
  author    = {Mahaut, Mat{\'e}o and
               Aina, Laura and
               Czarnowska, Paula and
               Hardalov, Momchil and
               M{\"u}ller, Thomas and
               Marquez, Lluis},
  booktitle = {Proceedings of the 62nd Annual Meeting of the Association for
               Computational Linguistics (Volume 1: Long Papers)},
  pages     = {4554--4570},
  year      = {2024},
  publisher = {Association for Computational Linguistics},
  doi       = {10.18653/v1/2024.acl-long.250}
}

@inproceedings{beigi2024internalinspector,
  title     = {{InternalInspector} $I^2$: Robust Confidence Estimation in
               {LLM}s through Internal States},
  author    = {Beigi, Mohammad and
               Shen, Ying and
               Yang, Runing and
               Lin, Zihao and
               Wang, Qifan and
               Mohan, Ankith and
               He, Jianfeng and
               Jin, Ming and
               Lu, Chang-Tien and
               Huang, Lifu},
  booktitle = {Findings of the Association for Computational Linguistics:
               EMNLP 2024},
  pages     = {12847--12865},
  year      = {2024},
  publisher = {Association for Computational Linguistics},
  doi       = {10.18653/v1/2024.findings-emnlp.751}
}

@inproceedings{scialom2022continual,
  title     = {Fine-tuned Language Models are Continual Learners},
  author    = {Scialom, Thomas and
               Chakrabarty, Tuhin and
               Muresan, Smaranda},
  booktitle = {Proceedings of the 2022 Conference on Empirical Methods in
               Natural Language Processing},
  pages     = {6107--6122},
  year      = {2022},
  publisher = {Association for Computational Linguistics},
  doi       = {10.18653/v1/2022.emnlp-main.410}
}

@inproceedings{he2025kpig,
  title     = {Don't Half-listen: Capturing Key-part Information in Continual
               Instruction Tuning},
  author    = {He, Yongquan and
               Zhang, Wenyuan and
               Huang, Xuancheng and
               Zhang, Peng and
               Meng, Lingxun and
               Zhou, Xiang and
               Zeng, Ke and
               Cai, Xunliang},
  booktitle = {Proceedings of the 63rd Annual Meeting of the Association for
               Computational Linguistics (Volume 1: Long Papers)},
  pages     = {23649--23668},
  year      = {2025},
  publisher = {Association for Computational Linguistics},
  doi       = {10.18653/v1/2025.acl-long.1153}
}

@article{kadavath2022language,
  title   = {Language Models (Mostly) Know What They Know},
  author  = {Kadavath, Saurav and
             Conerly, Tom and
             Askell, Amanda and
             Henighan, Tom and
             Drain, Dawn and
             Perez, Ethan and
             Schiefer, Nicholas and
             Hatfield-Dodds, Zac and
             DasSarma, Nova and
             Tran-Johnson, Eli and
             Johnston, Scott and
             El-Showk, Sheer and
             Jones, Andy and
             Elhage, Nelson and
             Hume, Tristan and
             Chen, Anna and
             Bai, Yuntao and
             Bowman, Sam and
             Fort, Stanislav and
             Ganguli, Deep and
             Hernandez, Danny and
             Jacobson, Josh and
             Kernion, Jackson and
             Kravec, Shauna and
             Lovitt, Liane and
             Ndousse, Kamal and
             Olsson, Catherine and
             Ringer, Sam and
             Amodei, Dario and
             Brown, Tom and
             Clark, Jack and
             Joseph, Nicholas and
             Mann, Ben and
             McCandlish, Sam and
             Olah, Chris and
             Kaplan, Jared},
  journal = {arXiv preprint arXiv:2207.05221},
  year    = {2022},
  url     = {https://arxiv.org/abs/2207.05221}
}

@article{chan2023chateval,
  title   = {{ChatEval}: Towards Better {LLM}-based Evaluators through
             Multi-Agent Debate},
  author  = {Chan, Chi-Min and
             Chen, Weize and
             Su, Yusheng and
             Yu, Jianxuan and
             Xue, Wei and
             Zhang, Shanghang and
             Fu, Jie and
             Liu, Zhiyuan},
  journal = {arXiv preprint arXiv:2308.07201},
  year    = {2023},
  url     = {https://arxiv.org/abs/2308.07201}
}

@article{verga2024replacing,
  title   = {Replacing Judges with Juries: Evaluating {LLM} Generations with a
             Panel of Diverse Models},
  author  = {Verga, Pat and
             Hofstatter, Sebastian and
             Althammer, Sophia and
             Su, Yixuan and
             Piktus, Aleksandra and
             Arkhangorodsky, Arkady and
             Xu, Minjie and
             White, Naomi and
             Lewis, Patrick},
  journal = {arXiv preprint arXiv:2404.18796},
  year    = {2024},
  url     = {https://arxiv.org/abs/2404.18796}
}

@inproceedings{lewis2020retrieval,
  title     = {Retrieval-Augmented Generation for Knowledge-Intensive {NLP}
               Tasks},
  author    = {Lewis, Patrick and
               Perez, Ethan and
               Piktus, Aleksandra and
               Petroni, Fabio and
               Karpukhin, Vladimir and
               Goyal, Naman and
               K{\"u}ttler, Heinrich and
               Lewis, Mike and
               Yih, Wen-tau and
               Rockt{\"a}schel, Tim and
               Riedel, Sebastian and
               Kiela, Douwe},
  booktitle = {Advances in Neural Information Processing Systems},
  volume    = {33},
  year      = {2020},
  url       = {https://arxiv.org/abs/2005.11401}
}
\bibliographystyle{iclr2027_conference}

\newpage


\appendix

\section{Additional Details of \vera-Guided Periodic Adaptation}
\label{app:vera-algorithm}

We provide score-processing conventions, the full training objective, and the
adaptation procedure omitted from Section~\ref{sec:methodology}.
Layer-selection criteria and experiments are reported separately in
Appendix~\ref{app:layer-selection-analysis}.

\subsection{Reliability-Score Implementation}
\label{app:vera-calibration}

\paragraph{Direction estimation.}
We use the pre-verdict hidden activation specified in
Section~\ref{sec:vera-directions} for both construction and scoring.
The global comparison omits predicted-verdict conditioning:
\begin{equation}
    v_\ell^{\mathrm{global}}
    =\mathbb{E}[h_\ell(x)\mid r_\theta(x)=1]
     -\mathbb{E}[h_\ell(x)\mid r_\theta(x)=0].
    \label{eq:global-vera-axis-app}
\end{equation}
This is a comparison method, not an automatic fallback for small verdict
groups.
For normalization, including Eq.~\ref{eq:vera-refresh}, we use
\begin{equation}
    \operatorname{Normalize}(v)
    =\frac{v}{\|v\|_2+\epsilon},\qquad \epsilon=10^{-8}.
    \label{eq:app-vera-normalization}
\end{equation}
An exactly zero direction receives score zero.
On refresh rounds, both the new and stored directions are normalized before
mixing, and the weighted sum is normalized again with $\gamma=0.5$.

\paragraph{Calibration.}
We instantiate the verdict-specific map in Eq.~\ref{eq:vera-value} as
\begin{equation}
    C_c(s)=\sigma(a_cs+b_c),
    \label{eq:app-vera-calibration}
\end{equation}
where $\sigma$ is the logistic function and $(a_c,b_c)$ are fitted to
correctness labels.
Initial maps are fitted on held-out reference judgments.
A map is fitted when its verdict group contains at least 20 examples and both
correctness outcomes; otherwise, we use $(a_c,b_c)=(1,0)$, yielding
$C_c(s)=\sigma(s)$.
This default is a score-to-probability transformation, not a guarantee of
calibration.
On refresh rounds, directions and calibration maps are re-estimated using
post-update judgments from $\mathcal{D}_t$; the same batch is then used for
replay selection.
These label-informed scores guide future training rather than constitute an
independent evaluation of calibration quality.

\subsection{Full Training Objective}
\label{app:vera-objective}

For $\mathcal{S}_t=\mathcal{D}_t\cup\mathcal{M}_{t-1}$, let
$w_t(x)=1$ on current examples and $w_t(x)=w_r$ on replay examples, with
$\overline w_t=\mathbb{E}_{x\in\mathcal{S}_t}[w_t(x)]$.
The cross-entropy and KL terms in Eq.~\ref{eq:vera-loss} are
\begin{align}
    \mathcal{L}_{\mathrm{CE}}
    &=\mathbb{E}_{x\in\mathcal{S}_t}\!\left[
        -\frac{w_t(x)}{\overline w_t}\log\pi_\theta(y^\star\mid x)
    \right],
    \label{eq:vera-ce}\\
    \mathcal{L}_{\mathrm{KL}}
    &=\mathbb{E}_{x\in\mathcal{S}_t}\!\left[
        \beta_t(x)D_{\mathrm{KL}}\!\left(
        \pi_{\theta_{t-1}}(\cdot\mid x)
        \,\Vert\,\pi_\theta(\cdot\mid x)\right)
    \right].
    \label{eq:vera-kl}
\end{align}
The dataset-wide normalization preserves the relative replay weight while
keeping the mean cross-entropy weight at one.
We relax the KL penalty for incorrect current judgments:
\begin{equation}
    \beta_t(x)=
    \begin{cases}
        \beta_{\mathrm{wr}},
        & x\in\mathcal{D}_t\ \text{and}\ \hat y_t^-(x)\neq y^\star,\\
        \beta_{\mathrm{cor}}, & \text{otherwise}.
    \end{cases}
    \label{eq:app-beta}
\end{equation}
Thus, replay examples and previously correct current judgments receive the
stronger preservation weight.
The unlikelihood penalty in Eq.~\ref{eq:vera-ul} is
\begin{equation}
    u_t(x;\theta)
    =-\log\!\left(1-\min\{
        \pi_\theta(\hat y_t^-(x)\mid x),\kappa\}\right).
    \label{eq:vera-unlikelihood}
\end{equation}
The cap bounds the penalty; above $\kappa$, this term has zero gradient with
respect to the predicted probability.

We use $\alpha=4$, $\lambda=0.5$, $\kappa=0.99$,
$\beta_{\mathrm{cor}}=0.02$, and $\beta_{\mathrm{wr}}=0.002$.
The error weight $\rho(x)$ is three for missed \textsc{Critical} cases in
proprietary auditing or missed ties in public judging, and one otherwise.
Task-specific replay settings are reported with the experimental setup.

\subsection{Complete Periodic Adaptation Procedure}
\label{app:vera-pseudocode}

Algorithm~\ref{alg:vera-batch} summarizes the complete procedure.
Reliability ranks for corrective training are computed using the pre-update
judge $\theta_{t-1}$, whereas residual replay is selected using the updated
judge $\theta_t$ after any scheduled reliability refresh.

\begin{algorithm}[t]
\caption{\vera-Guided Periodic Adaptation}
\label{alg:vera-batch}
\small
\begin{algorithmic}[1]

\Require Initial judge $\theta_0$;
selected layer $\ell^\star$;
initial reliability directions $\{v_{c,0}\}_{c\in\mathcal{Y}}$;
initial calibration maps $\{C_{c,0}\}_{c\in\mathcal{Y}}$;
verified batches $\{\mathcal{D}_t\}_{t=1}^{T}$;
initial historical replay memory $\mathcal{M}_0$;
selection budgets $\{K_t\}_{t=1}^{T}$;
update hyperparameters
$\alpha,\lambda,\kappa,w_r,\rho,\beta_{\mathrm{cor}},\beta_{\mathrm{wr}}$;
refresh rate $\gamma$ and refresh schedule

\State $\mathcal{M}_0
\gets \operatorname{Dedup}_{\mathrm{uid}}(\mathcal{M}_0)$

\For{$t=1,\ldots,T$}

    \Statex
    \State \textbf{// Pre-update reliability estimation}

    \ForAll{$x\in\mathcal{D}_t$}
        \State Compute pre-update verdict
        $\hat y_t^-(x)\gets\hat y_{\theta_{t-1}}(x)$
        \State Extract $h_{\ell^\star}(x)$ using the frozen judge
        $\theta_{t-1}$
        \State Compute the \vera~score using
        $v_{\hat y_t^-(x),t-1}$ and Eq.~\ref{eq:vera-score}
    \EndFor

    \State Compute within-verdict ranks
    $\{q_t^-(x):x\in\mathcal{D}_t\}$
    using Eq.~\ref{eq:vera-rank}

    \State Compute normalized corrective gates
    $\{\widetilde g_t(x):x\in\mathcal{D}_t\}$
    using Eq.~\ref{eq:vera-gate}

    \Statex
    \State \textbf{// Periodic model update}

    \State $\mathcal{S}_t
    \gets \mathcal{D}_t\cup\mathcal{M}_{t-1}$

    \State Compute reference verdict distributions on $\mathcal{S}_t$
    using $\theta_{t-1}$

    \State Initialize $\theta\gets\theta_{t-1}$

    \State Optimize $\theta$ on $\mathcal{S}_t$ using
    Eq.~\ref{eq:vera-loss}, holding pre-update signals and weights fixed

    \State $\theta_t\gets\theta$

    \Statex
    \State \textbf{// Post-update reliability refresh}

    \State Run $\theta_t$ on $\mathcal{D}_t$ and cache updated verdicts,
    hidden activations, and realized correctness

    \If{$t$ is a reliability-refresh round}
        \State Estimate
        $\{v^{\mathrm{new}}_{c,t}\}_{c\in\mathcal{Y}}$
        using Eq.~\ref{eq:vera-axis}
        \State Normalize the new and previous directions and refresh
        $\{v_{c,t}\}_{c\in\mathcal{Y}}$
        using Eq.~\ref{eq:vera-refresh}
        \State Refit $\{C_{c,t}\}_{c\in\mathcal{Y}}$ on
        $\mathcal{D}_t$ using the updated judgments and refreshed directions
    \Else
        \State $v_{c,t}\gets v_{c,t-1}$ for all $c\in\mathcal{Y}$
        \State $C_{c,t}\gets C_{c,t-1}$ for all $c\in\mathcal{Y}$
    \EndIf

    \Statex
    \State \textbf{// Residual replay for the next round}

    \ForAll{$x\in\mathcal{D}_t$}
        \State Compute $V_t(x)$ using the updated verdict,
        cached activation, and current reliability estimator
        \State $\Delta_{\vera,t}(x)
        \gets |r_{\theta_t}(x)-V_t(x)|$
    \EndFor

    \State $\mathcal{C}_t
    \gets \operatorname{Top}_{K_t}
    (\mathcal{D}_t;\Delta_{\vera,t})$

    \State $\mathcal{M}_t
    \gets \operatorname{Dedup}_{\mathrm{uid}}
    (\mathcal{M}_{t-1}\cup\mathcal{C}_t)$

\EndFor

\State \Return
$\theta_T$,
$\{v_{c,T}\}_{c\in\mathcal{Y}}$,
$\{C_{c,T}\}_{c\in\mathcal{Y}}$,
and $\mathcal{M}_T$

\end{algorithmic}
\end{algorithm}

\section{{\color{revisionblue}Hyperparameter Settings}}
\label{app:hyperparameters}

{\color{revisionblue}
\paragraph{Training configurations.}
Table~\ref{tab:app-training-hyperparameters} summarizes the training
configurations used for both judging tasks and model scales.
The IFT checkpoint initializes both the DPO baseline and \vera-guided
periodic adaptation.
{\color{reviewconsistency}
The proprietary periodic-IFT controls at both scales instead use a maximum
sequence length of $2{,}048$, an effective batch size of $128$, and one epoch
per feedback batch, matching \vera's context length and batch size while
retaining the IFT learning rate of $10^{-4}$.}

All configurations use LoRA with rank $16$, scaling parameter
$\alpha_{\mathrm{LoRA}}=32$, and dropout $0.05$.
Adapters are applied to the attention projections
(\texttt{q\_proj}, \texttt{k\_proj}, \texttt{v\_proj}, and
\texttt{o\_proj}) and feed-forward projections
(\texttt{gate\_proj}, \texttt{up\_proj}, and \texttt{down\_proj}).
The DPO baseline uses $\beta_{\mathrm{DPO}}=0.1$, distinct from the KL
coefficients in the \vera~objective.
Single-seed experiments use seed $13$; the seed set for multi-seed
experiments is $13$, $17$, and $23$.
}

\begin{table}[t]
    \color{revisionblue}
    \centering
    \small
    \setlength{\tabcolsep}{6pt}
    \begin{tabular}{@{}lccc@{}}
        \toprule
        Hyperparameter & IFT & DPO
        & \shortstack{\vera-guided\\periodic update} \\
        \midrule
        Optimizer
        & AdamW & AdamW & AdamW \\
        Maximum sequence length (tokens)
        & $4{,}096$ & $2{,}048$ & $2{,}048$ \\
        Peak learning rate
        & $1\times10^{-4}$ & $5\times10^{-6}$ & $5\times10^{-5}$ \\
        Training epochs
        & $2$ & $1$ & $1$ per feedback batch \\
        Effective global batch size
        & $256$ & $128$ & $128$ \\
        Warmup ratio
        & $0.03$ & $0.05$ & $0.05$ \\
        Learning-rate schedule
        & Cosine & Cosine & Cosine \\
        Weight decay
        & $0$ & $0$ & $0$ \\
        Numerical precision
        & \texttt{bfloat16} & \texttt{bfloat16} & \texttt{bfloat16} \\
        Gradient checkpointing
        & Enabled & Enabled & Enabled \\
        \bottomrule
    \end{tabular}
    \caption{
    \textbf{Training hyperparameters shared across tasks and model scales.}
    IFT initializes the judge; DPO and \vera-guided adaptation are subsequent
    training alternatives.
    Effective global batch sizes include gradient accumulation across eight
    data-parallel workers.
    For periodic adaptation, the epoch count applies to each feedback-batch
    update.
    }
    \label{tab:app-training-hyperparameters}
\end{table}

{\color{revisionblue}
\paragraph{\vera~objective and reliability refresh.}
The corrective objective uses rank-gate strength $\alpha=4$,
unlikelihood weight $\lambda=0.5$, and probability cap $\kappa=0.99$.
The KL coefficient is $\beta_{\mathrm{wr}}=0.002$ for incorrect
current-batch judgments and $\beta_{\mathrm{cor}}=0.02$ for all
remaining examples, including replay examples.
The task-specific error weight $\rho(x)$ is $3$ for missed
\textsc{Critical} verdicts in proprietary auditing and missed ties in
public judging, and $1$ otherwise.
Replay examples receive cross-entropy weight $w_r=0.15$ before the
dataset-wide normalization defined in
Appendix~\ref{app:vera-objective}.
Reliability refresh uses direction-smoothing coefficient $\gamma=0.5$;
the refresh interval is $1$ batch, i.e., refresh occurs after each of the
feedback batches.

\paragraph{Task-specific settings.}
Table~\ref{tab:app-vera-hyperparameters} reports the adaptation schedule,
representation layer, and replay budget for each task and scale.
The layer is selected before adaptation and remains fixed across all
updates and direction refreshes.
Selection criteria and validation sweeps are reported in
Appendix~\ref{app:layer-selection-analysis}.

Proprietary replay uses the same selection fraction for the initial pool
and each feedback batch. Public replay uses separate initial and per-batch
budgets, reported as both item counts and equivalent selection fractions.
For public judging, each selected item contributes both response orders,
yielding two training examples.
Replay memory accumulates selected examples across batches, as specified in
Algorithm~\ref{alg:vera-batch}.
}

\begin{table}[t]
    \color{revisionblue}
    \centering
    \small
    \setlength{\tabcolsep}{6pt}
    \begin{tabular}{@{}lrrrr@{}}
        \toprule
        & \multicolumn{2}{c}{Proprietary}
        & \multicolumn{2}{c}{Public} \\
        \cmidrule(lr){2-3}\cmidrule(lr){4-5}
        Hyperparameter & 8B & 14B & 8B & 14B \\
        \midrule
        Feedback batches $T$
        & $4$ & $4$ & $5$ & $5$ \\
        Selected layer $\ell^\star$
        & $20$ & $38$ & $33$ & $32$ \\
        Initial selection fraction (\%)
        & $1.00$ & $1.50$ & $0.75$ & $1.50$ \\
        Feedback selection fraction (\%)
        & $1.00$ & $1.50$ & $3.83$ & $7.66$ \\
        Final replay examples
        & $747$ & $1{,}120$ & $3{,}880$ & $7{,}760$ \\
        \bottomrule
    \end{tabular}
    \caption{
    \textbf{Task- and scale-specific adaptation settings.}
    Layer indices are zero-indexed decoder-block indices.
    Proprietary selection uses fraction $f$ for both the initial pool and
    feedback batches.
    Public fractions are rounded percentages of $98{,}400$ unique source
    items and $6{,}268$ unique items per feedback batch, respectively;
    the listed item counts specify the exact budgets.
    Each public item contributes two response orders.
    Final public replay counts therefore equal
    $2(740+5\times240)=3{,}880$ at 8B and
    $2(1{,}480+5\times480)=7{,}760$ at 14B.
    }
    \label{tab:app-vera-hyperparameters}
\end{table}

{\color{revisionblue}
\paragraph{Replay-budget analysis.}
Appendix~\ref{app:internal-replay-budget} evaluates
$f\in\{0.010,0.015,0.020\}$ for proprietary replay while keeping the other
hyperparameters fixed, to examine sensitivity to replay volume.
The independent proprietary forward set is reserved for evaluation and is
not used for training, layer selection, or hyperparameter selection.

\paragraph{Periodic RAG baseline.}
The proprietary periodic-RAG baseline keeps the IFT judge fixed and retrieves
the three nearest labeled examples available before the current evaluation
period.
Retrieval uses normalized embeddings from a sentence-transformer encoder
(\texttt{intfloat/multilingual-e5-large}) and a FAISS
\texttt{IndexFlatIP} index.
Examples sharing the query identifier are excluded, and each retrieved input
is truncated to $1{,}500$ characters.

\paragraph{Computing environment.}
Training uses eight NVIDIA A100 GPUs with $40$\,GB of memory each,
with data parallelism and no model sharding.
The software environment comprises PyTorch~2.12.1,
Transformers~5.12.1, TRL~1.7.0, PEFT~0.19.1,
Accelerate~1.14.0, and Datasets~5.0.0.
}

\section{Additional Reliability Evaluations}
\label{app:additional-reliability}

\subsection{{\color{revisionblue}AUROC Convention and Reliability Estimators}}
\label{app:auroc-convention}

Throughout the paper, an unqualified AUROC is the correctness-ranking AUROC
of the verdict-conditioned \vera~score:
\begin{equation}
    \operatorname{AUROC}
    \left(
        s_{\vera}(x),
        \mathbf{1}\{\hat y_\theta(x)=y^\star\}
    \right).
\end{equation}
An AUROC based on another score is named explicitly as
\emph{softmax-confidence AUROC} or \emph{verbalized-confidence AUROC}.

{\color{revisionblue}
\paragraph{Compared hidden-state estimators.}
We compare three estimators using Qwen3-8B at the Frozen and IFT checkpoints
on both judging tasks.
The \emph{global} estimator applies the single difference-of-means direction
in Eq.~\ref{eq:global-vera-axis} to every judgment.
The \emph{per-class} estimator is \vera: it constructs a separate
difference-of-means direction for each predicted verdict using
Eq.~\ref{eq:vera-axis} and scores each judgment against the corresponding
direction.
Both constructions use sample means of labeled reference judgments.
The \emph{LR probe} fits a class-balanced logistic regression to predict
correctness from $\ell_2$-normalized activations at each layer, using its
decision value for AUROC.
The LR fit is capped at $20{,}000$ training judgments.

For each estimator, we select its layer on a disjoint validation split and
report correctness AUROC on the corresponding test set.
For proprietary judging, construction, layer selection, and evaluation use
the proprietary training, validation, and test splits, respectively.
For the external comparison, estimators are constructed on JudgeLM training
judgments, layers are selected on JudgeLM validation judgments, and transfer
is evaluated on MT-Bench-HJ.
MT-Bench-HJ labels are not used for estimator construction or layer selection.
}

\begin{table*}[t]
    \color{revisionblue}
    \centering
    \small
    \setlength{\tabcolsep}{6pt}
    \begin{tabular}{@{}lllrrr@{}}
        \toprule
        Setting & Model & Stage
        & Global & Per-class (\vera) & LR probe \\
        \midrule
        \multirow{2}{*}{Proprietary}
        & \multirow{2}{*}{Qwen3-8B}
        & Frozen & 0.671 & \textbf{0.812} & 0.761 \\
        & & IFT & 0.848 & \textbf{0.867} & 0.862 \\
        \midrule
        \multirow{2}{*}{MT-Bench-HJ}
        & \multirow{2}{*}{Qwen3-8B}
        & Frozen & $0.534$ & $\mathbf{0.712}$ & $0.669$ \\
        & & IFT & $\mathbf{0.730\pm0.011}$ & $0.719\pm0.009$ & $0.725\pm0.006$ \\
        \bottomrule
    \end{tabular}
    \caption{
    \textbf{Correctness AUROC for three hidden-state estimators at Qwen3-8B.}
    Frozen denotes the pre-IFT checkpoint.
    Each estimator uses its own validation-selected layer.
    Proprietary rows use the proprietary training, validation, and test splits;
    MT-Bench-HJ rows use JudgeLM for construction and layer selection and
    MT-Bench-HJ for transfer evaluation.
    MT-Bench-HJ ties are valid verdicts and are retained.
    Frozen proprietary abstentions are excluded consistently across estimators
    ($n=1{,}983$ retained judgments); the proprietary IFT judge has no
    abstentions.
    MT-Bench-HJ IFT entries report means and sample standard deviations over three seeds;
    Frozen is a single evaluation.
    {\color{reviewconsistency}
    MT-Bench-HJ results use regenerated predictions, as in
    Table~\ref{tab:app-posttraining}; Table~\ref{tab:main-reliability}
    retains the original reference evaluations.}
    Bold marks the highest AUROC or mean in each row.
    }
    \label{tab:app-hidden-state-estimators}
\end{table*}

{\color{revisionblue}
Per-class directions achieve the highest AUROC in three of the four
comparisons.
On the proprietary task, \vera~outperforms both alternatives at the Frozen
and IFT checkpoints.
For the Frozen external judge, it reaches $0.712$, compared with $0.534$
for the global direction and $0.669$ for the LR probe.
After JudgeLM IFT, however, the global direction exceeds per-class \vera~
by $0.011$ in mean AUROC ($0.730$ versus $0.719$).
Verdict conditioning is therefore beneficial in these comparisons but does
not uniformly yield the highest AUROC.

We use per-class \vera~to account for predicted-verdict composition without
fitting an additional classifier.
Its direction construction is training-free with respect to parameter
optimization: it requires labeled reference judgments, but neither
gradient-based probe optimization nor updates to the judge.
}

\subsection{{\color{revisionblue}Reliability Geometry Across Judging Distributions}}
\label{app:reliability-geometry}

{\color{revisionblue}
We visualize correctness-relevant geometry at the checkpoints used to initialize or benchmark
\vera-guided periodic updates.
All panels use Qwen3-8B and compare the Frozen, IFT, and DPO
checkpoints.
The proprietary visualization uses the fixed IFT-selected layer 20, and the
public visualizations use the JudgeLM-selected layer 33.
Within each displayed split and predicted-verdict group, the direction is
estimated with two-fold cross-fitting, and each held-out fold is scored with a
direction that excludes that row.
Colors show the resulting score standardized within the predicted-verdict
group; blue denotes a higher score and red a lower score.
Marker shapes identify the predicted verdict, and black circles mark the same
examples that all three checkpoints judge incorrectly.

For visualization only, each panel independently reduces the selected-layer
activations to 50 principal components and then applies t-SNE with perplexity
30 and seed 0.
Consequently, absolute positions and distances must not be compared across
panels.
The cross-fitted directions in these figures are diagnostic and use labels
from the displayed split; they are distinct from the source-axis protocol in
Table~\ref{tab:app-vera-transfer}, which transfers directions constructed on
JudgeLM without target labels.
}

\begin{figure*}[t]
    \color{revisionblue}
    \centering
    \includegraphics[width=\textwidth]
    {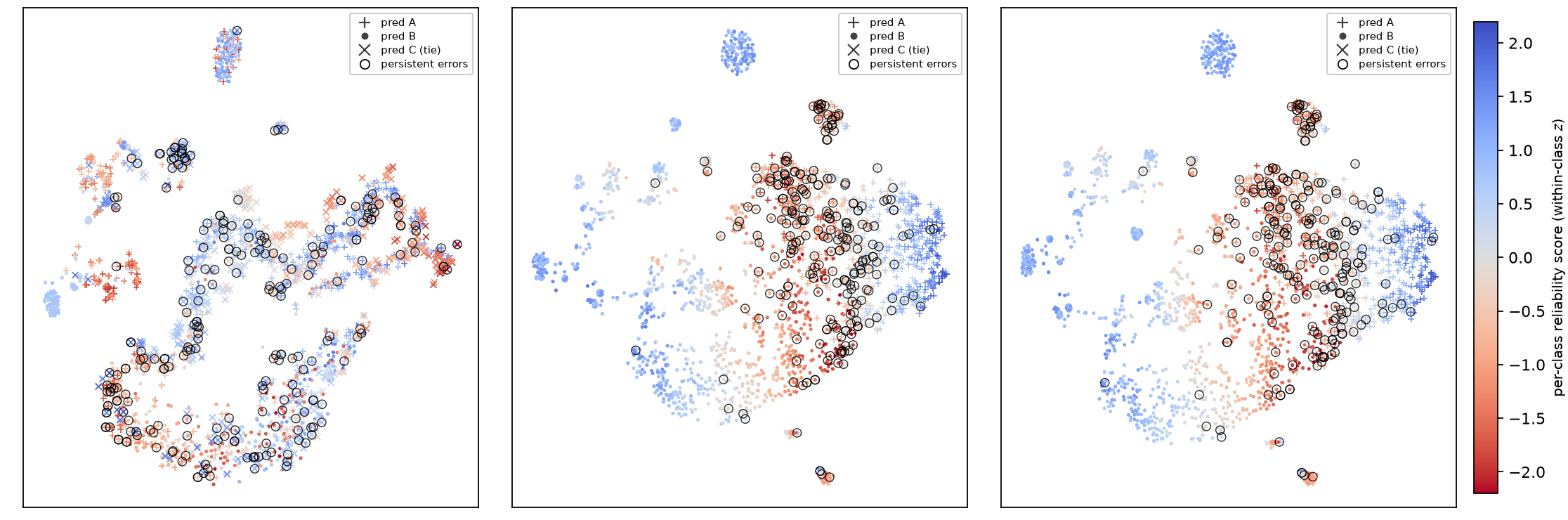}
    \vspace{-1mm}

    \begin{minipage}[t]{0.32\textwidth}
        \centering\small (a) Frozen
    \end{minipage}\hfill
    \begin{minipage}[t]{0.32\textwidth}
        \centering\small (b) Proprietary IFT
    \end{minipage}\hfill
    \begin{minipage}[t]{0.32\textwidth}
        \centering\small (c) Proprietary DPO
    \end{minipage}
    \caption{
    \textbf{Correctness-relevant geometry on the proprietary test set.}
    Each panel contains the same $2{,}634$ judgments at the fixed layer used
    for 8B periodic adaptation.
    After task-specific training, the score varies systematically within each
    predicted verdict, and the examples that remain incorrect at all three
    checkpoints occur predominantly in lower-score regions.
    This figure visualizes the reliability signal available to \vera; it is
    not an endpoint comparison of periodic-adaptation methods.
    }
    \label{fig:app-reliability-geometry-proprietary}
\end{figure*}

\begin{figure*}[t]
    \color{revisionblue}
    \centering
    {\small\textbf{JudgeLM validation}}\par
    \includegraphics[width=\textwidth]
    {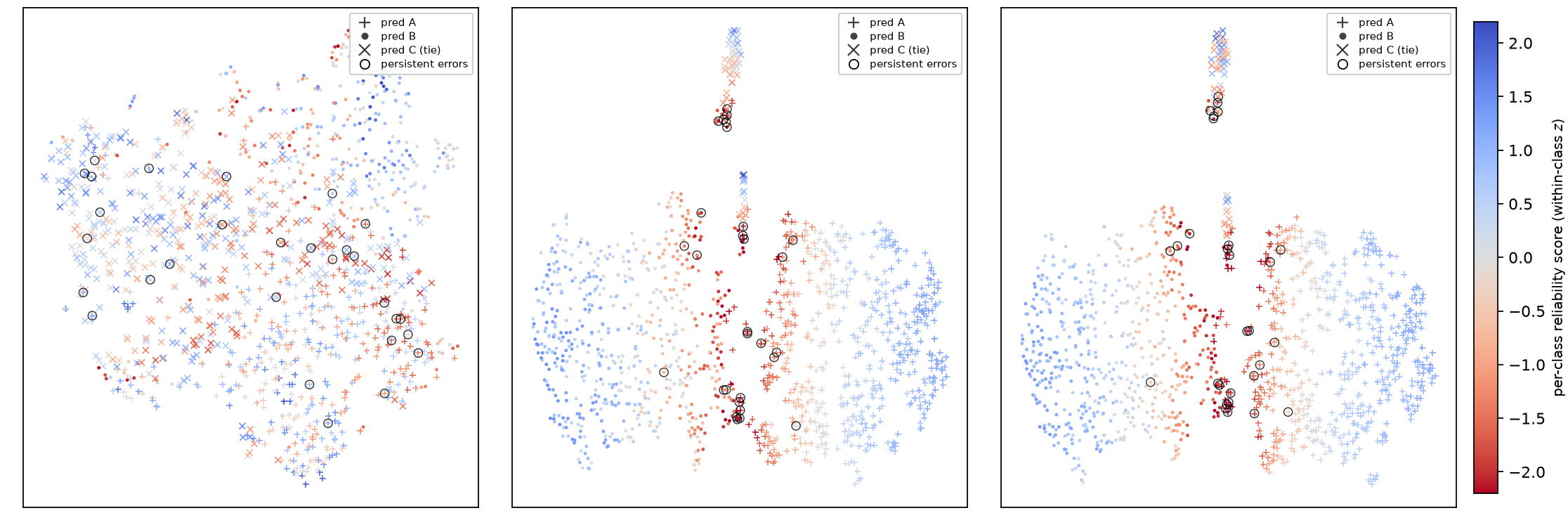}
    \vspace{-1mm}

    \begin{minipage}[t]{0.32\textwidth}
        \centering\small (a) Frozen
    \end{minipage}\hfill
    \begin{minipage}[t]{0.32\textwidth}
        \centering\small (b) JudgeLM IFT
    \end{minipage}\hfill
    \begin{minipage}[t]{0.32\textwidth}
        \centering\small (c) Arena DPO
    \end{minipage}

    \vspace{2mm}
    {\small\textbf{MT-Bench-HJ }}\par
    \includegraphics[width=\textwidth]
    {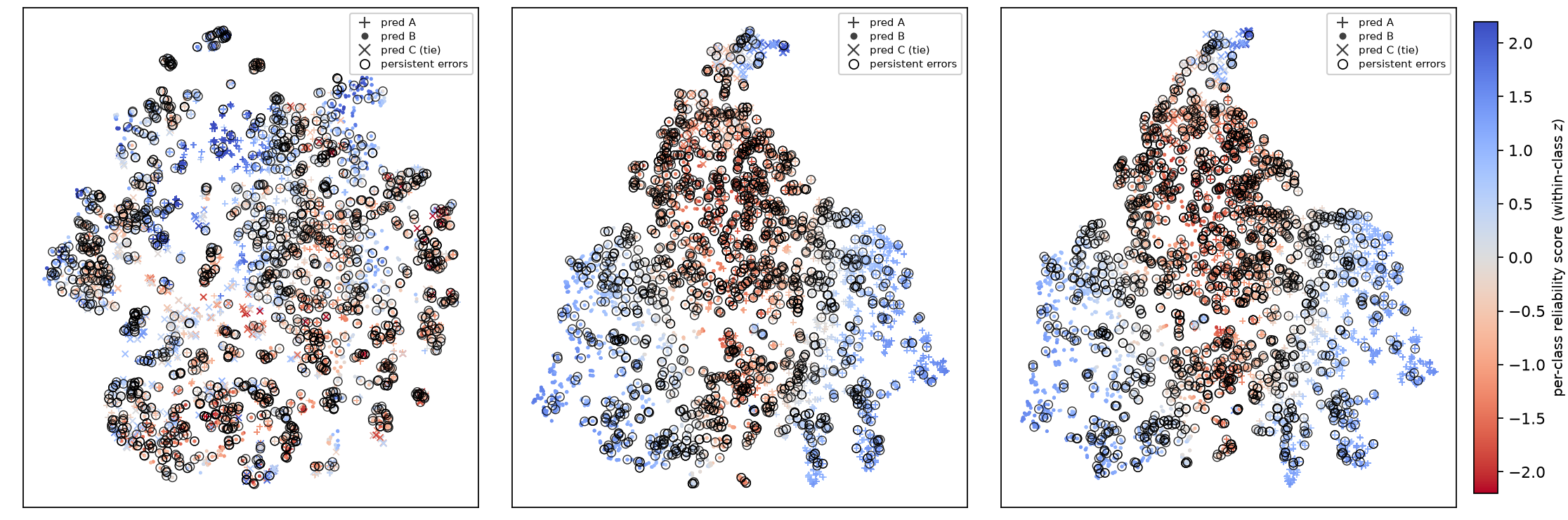}
    \vspace{-1mm}

    \begin{minipage}[t]{0.32\textwidth}
        \centering\small (d) Frozen
    \end{minipage}\hfill
    \begin{minipage}[t]{0.32\textwidth}
        \centering\small (e) JudgeLM IFT
    \end{minipage}\hfill
    \begin{minipage}[t]{0.32\textwidth}
        \centering\small (f) Arena DPO
    \end{minipage}
    \caption{
    \textbf{Correctness-relevant geometry on the public source and target
    distributions.}
    The top row shows $1{,}247$ JudgeLM validation judgments, the source
    distribution for IFT; the bottom row shows $6{,}710$ MT-Bench-HJ judgments,
    which are not used for judge training or layer selection.
    JudgeLM IFT leaves few source errors, so the top-row error geometry should
    be interpreted qualitatively.
    On MT-Bench-HJ, lower-score regions continue to contain many of the
    judgments that remain incorrect across checkpoints, showing that
    correctness-relevant structure remains present at the JudgeLM-selected
    layer under the target-distribution shift.
    }
    \label{fig:app-reliability-geometry-public}
\end{figure*}

{\color{revisionblue}
Figures~\ref{fig:app-reliability-geometry-proprietary}
and~\ref{fig:app-reliability-geometry-public} support the premise that a
verdict-conditioned score can distinguish difficult judgments that standard
IFT and DPO leave unresolved.
The MT-Bench-HJ visualization does not by itself establish why periodic
adaptation improves accuracy.
Rather, it shows that correctness-relevant structure remains available at the
public representation layer on the held-out target, while
Table~\ref{tab:app-vera-transfer} establishes source-axis transfer without
MT-Bench-HJ labels and Table~\ref{tab:main-cross-dataset} reports the $3.28\%$ relative gain over the
stronger 8B baseline.
}

\subsection{Post-Training Quality and Correctness Ranking}
\label{app:posttraining-calibration}

{\color{revisionblue}
Table~\ref{tab:app-posttraining} compares verdict quality and correctness
ranking before task-specific training, after IFT, and after DPO.
This analysis complements Table~\ref{tab:main-reliability}: here AUROC is
computed from softmax confidence rather than the \vera~score.
}


\begin{table*}[t]
    \color{revisionblue}
    \centering
    \small
    \setlength{\tabcolsep}{6pt}
    \begin{tabular}{lllrrr}
        \toprule
        Setting & Model & Stage
        & Acc. (\%) & Macro-F1 (\%) & Softmax AUROC \\
        \midrule
        \multirow{6}{*}{Proprietary}
        & \multirow{3}{*}{8B}
        & Frozen
        & 49.5 & 56.5 & 0.543 \\
        & & IFT
        & $\mathbf{86.3\pm0.2}$
        & $\mathbf{84.3\pm0.1}$
        & $\mathbf{0.861\pm0.005}$ \\
        & & DPO
        & $\mathbf{86.3\pm0.2}$
        & $\mathbf{84.3\pm0.2}$
        & $0.859\pm0.007$ \\
        \cmidrule(lr){2-6}
        & \multirow{3}{*}{14B}
        & Frozen
        & 30.4 & 24.4 & 0.630 \\
        & & IFT
        & \textbf{87.1} & \textbf{84.8} & 0.866 \\
        & & DPO
        & 86.9 & 84.6 & \textbf{0.870} \\
        \midrule
        \multirow{6}{*}{MT-Bench-HJ}
        & \multirow{3}{*}{8B}
        & Frozen & $56.48$ & $51.20$ & $0.640$ \\
        & & IFT & $64.44\pm0.15$ & $\mathbf{57.19\pm0.48}$ & $0.732\pm0.006$ \\
        & & DPO & $\mathbf{64.56\pm0.18}$ & $56.94\pm0.24$ & $\mathbf{0.734\pm0.004}$ \\
        \cmidrule(lr){2-6}
        & \multirow{3}{*}{14B}
        & Frozen & $61.22$ & $48.36$ & $0.713$ \\
        & & IFT & $64.10\pm0.26$ & $56.81\pm0.17$ & $\mathbf{0.734\pm0.001}$ \\
        & & DPO
        & $\mathbf{64.52\pm0.13}$ & $\mathbf{57.45\pm0.45}$ & $0.728\pm0.002$ \\
        \bottomrule
    \end{tabular}
    \caption{
    \textbf{Judge quality and correctness ranking across post-training stages.}
    Accuracy and macro-F1 are reported as percentages; AUROC uses softmax
    confidence.
    Frozen denotes the pre-IFT checkpoint.
    For MT-Bench-HJ, IFT and DPO correspond to JudgeLM instruction fine-tuning
    and subsequent Arena preference optimization, respectively.
    Entries with $\pm$ report means and sample standard deviations over three seeds.
    Entries without $\pm$ report single evaluations.
    MT-Bench-HJ metrics here use the predictions regenerated for reliability
    evaluation; Table~\ref{tab:main-cross-dataset} uses the original benchmark predictions.
    {\color{reviewconsistency}
    Proprietary 8B Frozen metrics retain abstentions as incorrect, whereas
    Table~\ref{tab:main-reliability} excludes them for that comparison.}
    Bold marks the highest reported score or mean within each
    setting--model combination, including ties at the displayed precision.
    }
    \label{tab:app-posttraining}
\end{table*}

{\color{revisionblue}
From IFT to DPO, accuracy and macro-F1 change by at most $0.79\%$, and softmax-confidence AUROC changes by at most {\color{reviewconsistency}$0.006$} across the
two tasks and model scales at the displayed precision.
Thus, DPO does not consistently improve the reported verdict quality or
correctness ranking over IFT.
}

\subsection{Layer Selection Analysis}
\label{app:reliability-estimators}
\label{app:layer-selection-analysis}

{\color{revisionblue}
We select the representation layer by maximizing correctness separation on a
disjoint validation split, following Eq.~\ref{eq:vera-layer}.
At each candidate decoder-block output, verdict-conditioned directions are
estimated on training judgments and held fixed during validation.
The selection criterion is
\begin{equation}
    \mathcal{J}_{\mathrm{val}}(\ell)
    =
    \begin{cases}
        \operatorname{AUROC}_{\mathrm{val}}
        \!\left(s_{\vera,\ell},r_\theta\right),
        & \text{proprietary judging},\\[2mm]
        \mu_{\ell,1}^{\mathrm{val}}
        -
        \mu_{\ell,0}^{\mathrm{val}},
        & \text{public judging},
    \end{cases}
    \label{eq:app-layer-criterion}
\end{equation}
where $\mu_{\ell,b}^{\mathrm{val}}$ is the empirical mean within-verdict
rank among validation judgments with correctness $b\in\{0,1\}$:
\begin{equation}
\begin{aligned}
    q_\ell^{\mathrm{val}}(x)
    &=
    \widehat F_{\ell,\hat y_\theta(x)}^{\mathrm{val}}
    \!\left(s_{\vera,\ell}(x)\right),\\
    \mu_{\ell,b}^{\mathrm{val}}
    &=
    \mathbb{E}_{\mathrm{val}}
    \!\left[q_\ell^{\mathrm{val}}(x)\mid r_\theta(x)=b\right].
\end{aligned}
    \label{eq:app-layer-validation-rank}
\end{equation}
Here $\widehat F_{\ell,c}^{\mathrm{val}}$ is the empirical CDF of
validation scores within predicted-verdict group $c$ at layer $\ell$.
The proprietary criterion measures discrimination by the raw score, whereas
the public criterion measures separation after the normalization used for
corrective weighting.
Both criteria use validation judgments only; test labels do not enter
direction construction or layer selection.
All layer indices below are zero-indexed decoder-block indices.

\paragraph{Proprietary judging.}
Figure~\ref{fig:app-proprietary-layer-selection} reports the raw-score
validation AUROC sweeps for the proprietary setting.
The Frozen curves characterize the pre-IFT judges, while the IFT curves
determine the layers used for periodic adaptation.

\begin{figure*}[t]
    \centering
    \begin{minipage}[t]{0.49\textwidth}
        \centering
        \includegraphics[width=\linewidth]
        {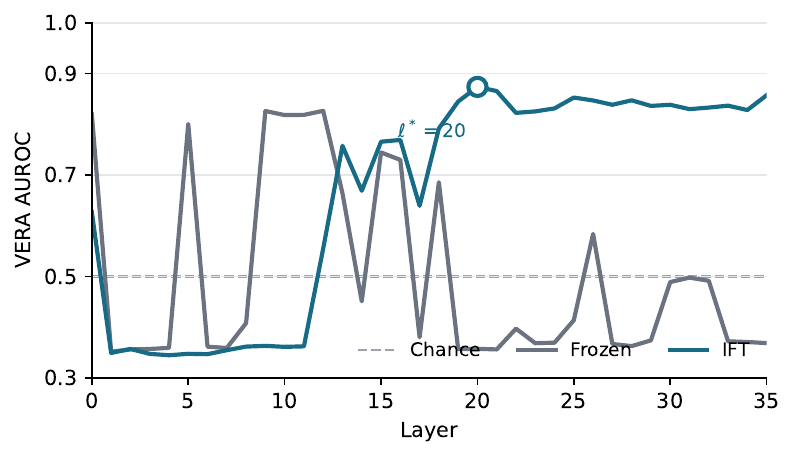}
        \small (a) Qwen3-8B
    \end{minipage}\hfill
    \begin{minipage}[t]{0.49\textwidth}
        \centering
        \includegraphics[width=\linewidth]
        {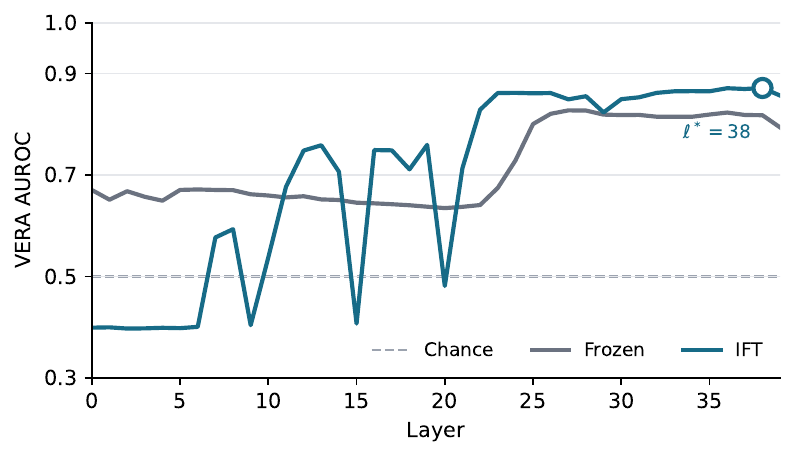}
        \small (b) Qwen3-14B
    \end{minipage}
    \caption{{\color{revisionblue}
    \textbf{Validation raw-score AUROC by layer on the proprietary task.}
    Directions are estimated on the training split and evaluated on the
    disjoint validation split.
    Open circles mark the IFT-selected layers used throughout periodic
    adaptation; Frozen curves are diagnostic.
    The dashed line denotes chance-level AUROC.}}
    \label{fig:app-proprietary-layer-selection}
\end{figure*}

The Frozen sweeps peak at layer 12 for Qwen3-8B and layer 27 for
Qwen3-14B, whereas the IFT sweeps peak at layers 20 and 38, respectively.
Because periodic adaptation starts from the IFT judge, we fix layers 20
and 38 throughout adaptation and subsequent direction refreshes.
The Frozen-selected layers are used only for the corresponding
checkpoint-level reliability analysis.

\paragraph{Public judging.}
For public adaptation, directions are estimated on JudgeLM training
judgments, and layer selection uses a disjoint JudgeLM validation split.
We maximize the correct--incorrect mean-rank difference in
Eq.~\ref{eq:app-layer-criterion}, matching the within-verdict ranks used
by the corrective objective.
This selects layer 33 at Qwen3-8B and layer 32 at Qwen3-14B.

For comparison with the proprietary sweeps,
Figure~\ref{fig:app-public-layer-selection} displays correctness AUROC
computed from the rank-normalized scores
$q_\ell^{\mathrm{val}}$, rather than the raw scores
$s_{\vera,\ell}$.
In the reported IFT sweeps, the selected layers also maximize this
rank-normalized AUROC.
However, the selection rule remains the mean-rank difference, not
maximization of raw-score AUROC.

\begin{figure*}[t]
    \centering
    \begin{minipage}[t]{0.49\textwidth}
        \centering
        \includegraphics[width=\linewidth]
        {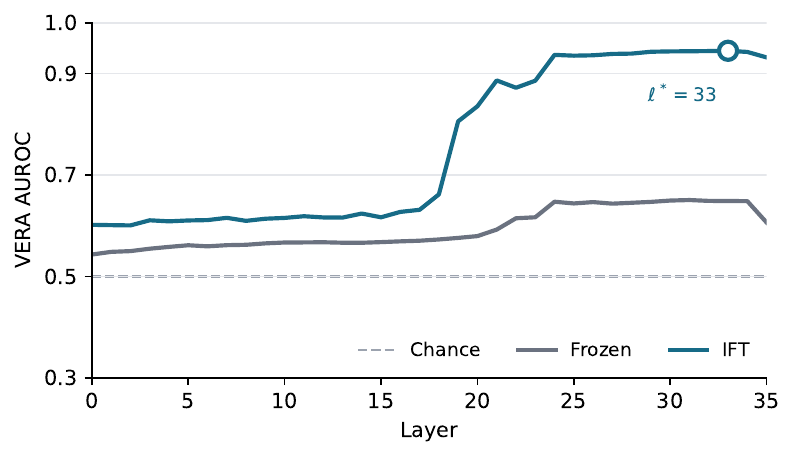}
        \small (a) Qwen3-8B
    \end{minipage}\hfill
    \begin{minipage}[t]{0.49\textwidth}
        \centering
        \includegraphics[width=\linewidth]
        {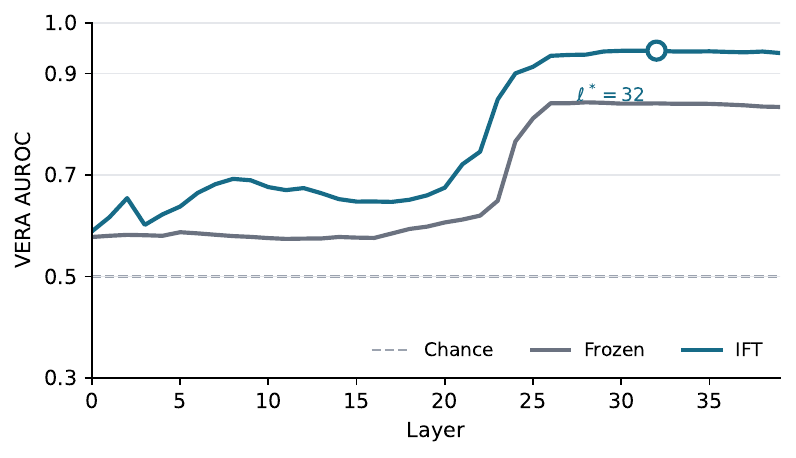}
        \small (b) Qwen3-14B
    \end{minipage}
    \caption{{\color{revisionblue}
    \textbf{Validation rank-normalized AUROC by layer on JudgeLM.}
    Directions are estimated on training judgments; validation scores are
    ranked within each predicted-verdict group before computing AUROC.
    Open circles mark the IFT layers selected by the correct--incorrect
    mean-rank gap, which also maximize rank-normalized AUROC in these sweeps.
    Frozen curves are diagnostic, and the dashed line denotes chance-level
    AUROC.}}
    \label{fig:app-public-layer-selection}
\end{figure*}

At the selected IFT layers, raw-score correctness AUROCs for predicted
verdicts A, B, and C are $0.948$, $0.950$, and $0.917$ at 8B, respectively.
The corresponding AUROCs at 14B are $0.948$, $0.970$, and $0.870$, respectively.
Thus, the shared layer retains substantial correctness discrimination
within all three verdict groups.
We use one layer across verdicts to avoid separate layer tuning for the
predicted-C group, which contains only approximately 50 validation examples
at each scale.
The selected IFT layer remains fixed throughout periodic adaptation.

{\color{proofreadmagenta}
\paragraph{Fixed layer and representation drift.}
The overall design pursues both accurate judgments and strong reliability
discrimination, assessed by correctness AUROC.
Initial validation selects a layer with strong correctness separation under
Eq.~\ref{eq:app-layer-criterion}; periodic direction refresh then tracks
changes in the judge's representations at that layer
(Eq.~\ref{eq:vera-refresh}).
Keeping $\ell^\star$ fixed provides a consistent representation for
reliability-guided correction and replay, while allowing the directions to
adapt with the judge.
Figures~\ref{fig:app-proprietary-layer-selection}
and~\ref{fig:app-public-layer-selection} establish the initial layer choice;
tracking changes in the optimal layer during adaptation is a separate
sensitivity question.
}

\paragraph{Separate Safety replay analysis.}
The RewardBench~2 Safety replay analysis selects its layer by grouped
cross-validation within the adaptation split.
It does not use the held-out Safety evaluation set or determine the layers
used in the main judging experiments.
}

\subsection{{\color{revisionblue}Source and In-Domain Axes for New Tasks}}
\label{app:vera-transfer}

{\color{revisionblue}
We compare source-transferred and in-domain reliability directions while
holding the Qwen3-8B JudgeLM-IFT checkpoint fixed.
Source \vera~constructs its directions and selects its layer on JudgeLM,
then transfers them unchanged to the target task.
In-domain \vera~instead uses verified target judgments to construct
verdict-conditioned directions and select the representation layer.
This comparison examines how access to target labels affects reliability
estimation, rather than changes to the judge itself.
}

\begin{table*}[t]
    \color{revisionblue}
    \centering
    \small
    \setlength{\tabcolsep}{5pt}
    \begin{tabular}{@{}lrrrrrr@{}}
        \toprule
        Signal & MT-Bench-HJ & Auto-J & JudgeBench
        & LLMBar & RewardBench 2 & Avg. \\
        \midrule
        Source \vera
        & \textbf{0.732} & 0.711 & 0.616
        & 0.565 & 0.760 & 0.677 \\
        In-domain \vera (ours)
        & \textbf{0.732} & \textbf{0.721} & \textbf{0.633}
        & \textbf{0.745} & \textbf{0.814} & \textbf{0.729} \\
        Softmax
        & \textbf{0.732} & 0.719 & 0.623
        & 0.576 & 0.796 & 0.689 \\
        Verbalized
        & 0.535 & 0.568 & 0.556
        & 0.466 & 0.545 & 0.534 \\
        \bottomrule
    \end{tabular}
    \caption{
    \textbf{Source and in-domain reliability directions at Qwen3-8B.}
    All entries report correctness-ranking AUROC for the same JudgeLM-IFT
    checkpoint.
    Source \vera~uses JudgeLM for direction construction and layer selection.
    In-domain \vera~uses grouped five-fold cross-validation on each target:
    examples sharing a \texttt{uid} remain in the same fold, directions are
    estimated on four folds, and scores are computed on the remaining fold.
    The reported in-domain result is selected by the highest pooled
    cross-validation AUROC across layers.
    On RewardBench~2, Source \vera~uses a global direction because JudgeLM
    lacks a fourth response option; in-domain \vera~constructs directions
    for all four options.
    Avg.\ is the unweighted mean across the five benchmarks.
    Bold marks the highest value in each column, including ties at the
    displayed precision.
    }
    \label{tab:app-vera-transfer}
\end{table*}

{\color{revisionblue}
The source and in-domain results differ by at most $0.017$ on MT-Bench-HJ,
Auto-J, and JudgeBench.
Larger gaps appear on LLMBar ($0.180$) and RewardBench~2 ($0.054$),
indicating that target judgments can reveal reliability information not
captured by the transferred source directions.
On RewardBench~2, in-domain \vera~reaches $0.814$, compared with $0.760$
for Source \vera~and $0.796$ for softmax.
{\color{proofreadmagenta}
Source \vera~retains correctness-ranking information across all five
unseen judging tasks, with AUROCs from $0.565$ to $0.760$ and a mean of
$0.677$.
Using verified target judgments to initialize task-specific directions and
select their layer raises the mean to $0.729$.
Relative to softmax, this target-calibrated estimator improves AUROC by
$0.169$ on LLMBar and $0.018$ on RewardBench~2, with a mean improvement of
$0.040$ across the five tasks.
}
However, the comparison changes both the construction data and, for
RewardBench~2, the global versus verdict-conditioned representation; it does
not isolate the effect of adding the fourth verdict direction.

\paragraph{Use on a new downstream task.}
When target labels are unavailable, Source \vera~provides a transferable
reliability estimate.
Transfer across the evaluated judging formats and domains shows that the
judge's internal representations retain correctness-relevant information
that generalizes beyond the source task.
When verified target judgments become available, they can be divided into
a construction split for estimating directions and a disjoint validation
split for selecting the layer.
The resulting estimator is then fixed before evaluating new judgments.

The in-domain entries in Table~\ref{tab:app-vera-transfer} are
target-label-informed reference estimates, not independently held-out
deployment results.
Although each judgment is scored by directions fitted on other folds,
selecting the best layer from the pooled cross-validation scores can introduce
selection optimism.
An independent target test set is therefore needed to evaluate the selected
estimator's generalization.
}

\section{Additional Periodic-Adaptation Results}
\label{app:additional-adaptation}

\subsection{Internal Replay Selection}
\label{app:internal-replay-selection}

We first distinguish the benefit of replay from that of its selection rule.
Table~\ref{tab:app-internal-replay-selection} compares no replay, random
replay, softmax-residual replay, and \vera-residual replay with Qwen3-8B
on the proprietary task.
Softmax residual replaces the reliability score $V$ with the predicted
verdict's softmax confidence in $|r-V|$.
The three replay selectors use matched replay budgets at $f=0.010$,
replay dose, update schedule, optimizer, and training seed.

\begin{table}[t]
    \centering
    \small
    \setlength{\tabcolsep}{6pt}
    \begin{tabular}{llrrr}
        \toprule
        Model & Selection rule & Anchor Rec. (\%)
        & Forward Rec. (\%) & Forward F1 (\%) \\
        \midrule
        \multirow{4}{*}{Qwen3-8B}
        & No replay & 78.7 & 61.6 & 63.9 \\
        & Random & 85.0 & 69.7 & 63.0 \\
        & Softmax residual & 85.5 & 69.7 & 63.0 \\
        & \vera~residual & \textbf{86.0} & \textbf{72.7} & \textbf{66.1} \\
        \bottomrule
    \end{tabular}
    \caption{
    \textbf{Replay-selection ablation on the proprietary task (\%).}
    Recall and F1 refer to the \textsc{Critical} verdict.
    All configurations use Qwen3-8B; replay selectors use matched budgets
    at $f=0.010$.
    Bold marks the best result in each column.
    }
    \label{tab:app-internal-replay-selection}
\end{table}

All three replay rules increase anchor and forward recall over no replay.
\vera~achieves higher forward recall and F1 than both random and
softmax-residual replay: $72.7\%$ versus $69.7\%$ recall and $66.1\%$ versus
$63.0\%$ F1.
Softmax-residual and random replay have identical forward metrics at the
reported precision.
These observations concern the 8B configuration.

\subsection{Replay-Budget Sensitivity}
\label{app:internal-replay-budget}

Table~\ref{tab:app-internal-replay-budget} varies the selected residual
fraction while keeping the update objective fixed.
Each row reports the final checkpoint after four periodic updates, evaluated
on the same earlier anchor and later forward sets.

\begin{table}[t]
    \centering
    \small
    \setlength{\tabcolsep}{6pt}
    \begin{tabular}{lcrrrr}
        \toprule
        Model & Fraction $f$
        & \multicolumn{2}{c}{Earlier anchor}
        & \multicolumn{2}{c}{Forward set} \\
        \cmidrule(lr){3-4}\cmidrule(lr){5-6}
        & & Recall (\%) & F1 (\%) & Recall (\%) & F1 (\%) \\
        \midrule
        \multirow{3}{*}{Qwen3-8B}
        & $0.010$ & 86.0 & 79.1 & 72.7 & \textbf{66.1} \\
        & $0.015$ & 85.3 & \textbf{79.5} & 69.7 & 65.1 \\
        & $0.020$ & \textbf{89.0} & 78.1 & \textbf{75.8} & 62.8 \\
        \midrule
        \multirow{3}{*}{Qwen3-14B}
        & $0.010$ & 84.2 & \textbf{80.0} & 72.7 & 68.6 \\
        & $0.015$ & 85.3 & 79.1 & 75.8 & \textbf{69.1} \\
        & $0.020$ & \textbf{88.4} & 79.2 & \textbf{77.8} & 67.2 \\
        \bottomrule
    \end{tabular}
    \caption{
    \textbf{Sensitivity to the reliability-residual replay budget (\%).}
    Metrics refer to the \textsc{Critical} verdict after four updates.
    The anchor and forward sets contain $2{,}634$ and $894$ examples,
    respectively. Each configuration uses one training seed.
    Bold marks the best result within each model scale and column.
    }
    \label{tab:app-internal-replay-budget}
\end{table}

{\color{proofreadmagenta}
Among the evaluated fractions, forward F1 is highest at $f=0.010$ for 8B
and $f=0.015$ for 14B.
}
At 14B, forward recall increases across the tested fractions, whereas F1
peaks at $f=0.015$.
At 8B, forward recall is non-monotonic, and $f=0.020$ gives the highest
recall but the lowest F1.
The replay fraction therefore changes the recall--F1 trade-off rather than
uniformly improving performance.
{\color{proofreadmagenta}
This ablation evaluates replay-budget sensitivity on the same independent
anchor and forward test sets, containing $2{,}634$ and $894$ judgments,
respectively. The sweep reports the effect of varying $f$.
}

\subsection{Internal Component Ablation}
\label{app:internal-components}

We remove replay, reliability refresh, or reliability-ranked unlikelihood
from the complete Qwen3-8B configuration while keeping the other components
fixed. Table~\ref{tab:app-internal-components} reports both anchor and
forward performance.

\begin{table}[t]
    \centering
    \small
    \setlength{\tabcolsep}{5pt}
    \begin{tabular}{lrrrrrr}
        \toprule
        Configuration
        & \multicolumn{3}{c}{Earlier anchor}
        & \multicolumn{3}{c}{Forward set} \\
        \cmidrule(lr){2-4}\cmidrule(lr){5-7}
        & Bal. Acc. & Recall & F1 & Bal. Acc. & Recall & F1 \\
        \midrule
        Complete \vera
        & 86.5 & \textbf{86.0} & 79.1
        & \textbf{83.4} & \textbf{72.7} & \textbf{66.1} \\
        $-$ replay
        & 85.2 & 78.7 & 79.2 & 78.8 & 61.6 & 63.9 \\
        $-$ reliability refresh
        & \textbf{86.9} & 84.9 & \textbf{80.3}
        & 81.2 & 67.7 & 64.7 \\
        $-$ rank-UL
        & 85.4 & 79.1 & 79.4 & 77.6 & 58.6 & 63.0 \\
        \bottomrule
    \end{tabular}
    \caption{
    \textbf{Component ablation on the proprietary task (\%).}
    All configurations use Qwen3-8B, $f=0.010$, and one training seed.
    Recall and F1 refer to the \textsc{Critical} verdict; balanced accuracy
    averages recall across both verdicts.
    Each ablation removes one component from the complete framework.
    Bold marks the highest score in each column.
    }
    \label{tab:app-internal-components}
\end{table}

Removing rank-UL, replay, and reliability refresh reduces forward recall by
$14.1\%$, $11.1\%$, and $5.0\%$, respectively.
The complete configuration also has the highest forward balanced accuracy
and F1. Retention effects differ: removing refresh slightly improves anchor
balanced accuracy and F1, while reducing anchor recall.
These comparisons support contributions to forward adaptation in the tested
configuration, rather than uniform improvements from every component on
every metric.

\subsection{RewardBench 2 Safety Selection and Repair}
\label{app:rb2-safety-retrieval}

We examine whether reliability-residual selection can help repair Safety
performance degraded during periodic adaptation.
The experiment separates identifying relevant examples from measuring the
effect of training on those examples.

\paragraph{Item-disjoint selection.}
RewardBench~2 contains $1{,}865$ items in the evaluated collection, each
scored under four cyclic response rotations.
We stratify by benchmark subset into $1{,}119$ selection items and $746$
evaluation items, grouping by $(\textsc{subset},\textsc{uid})$ so that
all rotations of an item remain in the same split.
The selection target consists of Safety items whose reference response
contains an explicit refusal or inability phrase and on which the adapted
judge makes an error.
Each selector ranks the full selection pool rather than a Safety-filtered
subset. {\color{proofreadmagenta}
For \vera, directions and calibration maps are fitted without Safety
judgments and then used to score the selection pool.
}
We select $300$ items and convert each into three pairwise examples,
producing a $900$-example repair set.
{\color{proofreadmagenta}
A target failure is a Safety item with a refusal or inability match in the
first $600$ characters of its reference response and an incorrect judgment
under at least one of its four rotations.
The \vera~item score is the arithmetic mean of the four rotation-level
residuals $|r-V|$; unavailable residuals receive $-1$ before averaging.
Items are ranked by decreasing score, with ties broken by
$(\textsc{subset},\textsc{uid})$.
}

\begin{table}[t]
    \centering
    \small
    \setlength{\tabcolsep}{6pt}
    \begin{tabular}{llrr}
        \toprule
        Model & Selector & Target-failure recall (\%) & Enrichment \\
        \midrule
        \multirow{2}{*}{Qwen3-8B}
        & Random & 28.1 & $1.05\times$ \\
        & \vera~residual & \textbf{68.5} & $\mathbf{2.56\times}$ \\
        \midrule
        \multirow{2}{*}{Qwen3-14B}
        & Random & 26.3 & $0.98\times$ \\
        & \vera~residual & \textbf{81.6} & $\mathbf{3.04\times}$ \\
        \bottomrule
    \end{tabular}
    \caption{
    \textbf{Selection of RewardBench 2 Safety failures.}
    Each selector retains $300$ of the same $1{,}119$ selection items.
    Recall is the fraction of target failures selected; enrichment is their
    prevalence among selected items divided by their prevalence in the full
    selection pool ($8.0\%$ at 8B and $6.8\%$ at 14B).
    These metrics are computed on the selection split before the repair update.
    Bold marks the larger value within each model scale and column.
    }
    \label{tab:app-rb2-safety-retrieval}
\end{table}

\vera-residual selection retrieves a larger fraction of the target Safety
failures than random selection at both scales, despite ranking a pool that
contains all benchmark domains.
This establishes the selection result; whether the selected examples improve
subsequent judgments is tested separately below.

{\color{reviewretention}
\paragraph{Failure diagnostics.}
Refusal-stratified diagnostics for the repair experiment show that the
unrepaired checkpoints reverse the IFT judge's relative advantage on Safety
items whose preferred response contains a refusal or inability phrase.
This identifies appropriate-refusal judgments as a target for repair.
One possible hypothesis is that repeated exposure to difficult tie cases
drives this change.
}

\paragraph{Held-out repair.}
Starting from the same unrepaired post-stream checkpoint, we perform $24$
optimizer steps on either the random or \vera-selected repair set alone.
The optimizer, schedule, selected-item count, and training seed are matched
between selectors.
We evaluate Safety on the $180$ Safety items in the $746$-item evaluation
split and measure MT-Bench-HJ accuracy to assess the effect on preference
judging.

\begin{table}[t]
    \centering
    \small
    \setlength{\tabcolsep}{6pt}
    \begin{tabular}{llrrr}
        \toprule
        Model & Intervention & Safety Acc. (\%)
        & Safety gain (\%) & MT-Bench-HJ Acc. (\%) \\
        \midrule
        \multirow{3}{*}{Qwen3-8B}
        & Before repair & 49.6 & -- & \textbf{66.4} \\
        & Random replay & 80.6 & $+31.0$ & 65.7 \\
        & \vera~replay & \textbf{86.9} & $\mathbf{+37.3}$ & 65.2 \\
        \midrule
        \multirow{3}{*}{Qwen3-14B}
        & Before repair & 54.0 & -- & \textbf{68.0} \\
        & Random replay & 72.1 & $+18.1$ & 65.9 \\
        & \vera~replay & \textbf{78.6} & $\mathbf{+24.6}$ & 66.0 \\
        \bottomrule
    \end{tabular}
    \caption{
    \textbf{Held-out effect of Safety repair.}
    Both replay interventions use $900$ pairwise examples and $24$ optimizer
    steps. Safety gain is the absolute change (\%) from the
    unrepaired checkpoint at the same model scale.
    Selection and evaluation items are disjoint, with all rotations grouped
    by item. Bold marks the best value within each model scale and column.
    }
    \label{tab:app-rb2-safety-repair}
\end{table}

Relative to matched random replay, \vera~improves held-out Safety accuracy
by $6.3\%$  at 8B and $6.5\%$ at 14B.
Random replay also produces substantial recovery, so the selector's added
benefit should be distinguished from the effect of additional training.
Relative to the unrepaired checkpoints, \vera~repair reduces MT-Bench-HJ
accuracy by $1.2\%$ and $2.0\%$, respectively.
The experiment therefore supports the usefulness of reliability-residual
selection for targeted repair, while exposing a trade-off between restoring
Safety judgments and retaining preference-judging performance.

\section{Inference-Time Harnesses and Verbalized Reliability}
\label{app:inference-scaffolds}

\subsection{Configurations and Motivation}
\label{app:scaffold-configurations}

We ask whether changing the inference harness can improve the correctness
ranking induced by verbalized confidence while leaving model parameters
fixed. These training-free comparisons complement the internal-reliability
analysis: stronger judging performance or additional model calls need not
imply that the confidence reported by the final judge better separates
correct and incorrect verdicts.

We evaluate three harnesses: \textbf{0-judge}, a single direct judge call;
\textbf{3-subjudge}, three independent judgments aggregated by a
meta-judge; and \textbf{3-subjudge + RAG}, which additionally supplies
retrieved labeled examples to the meta-judge.
The panel construction is motivated by multi-judge
evaluation~\citep{chan2023chateval,verga2024replacing}, and retrieval
follows the use of exemplars in retrieval-augmented
generation~\citep{lewis2020retrieval}.
These references motivate the configurations rather than specify identical
implementations. Prompts, temperatures, retrieval corpora, retrieval depths,
and panel size are matched across model scales for each harness.

\paragraph{Evaluation population.}
All harness AUROCs in this section use verbalized confidence, not the
\vera~score.
For the proprietary task, \textsc{Unsure} is not a reference verdict.
{\color{proofreadmagenta}
All-sample accuracy counts these abstentions as incorrect.
The reported AUROCs exclude \textsc{Unsure} before extracting confidence
and therefore use harness-dependent committed subsets.
The scorer reads \texttt{confidence.label}, accepting a scalar
\texttt{confidence} field as a fallback; values that cannot be converted
to a number receive $0.5$, and missing or unreadable prediction files are
excluded.
}
For MT-Bench-HJ, a tie is a valid verdict, not an abstention, and correctness
is determined by agreement with the three-way reference label.

\subsection{Proprietary Harness Results}
\label{app:internal-scaffolds}


Table~\ref{tab:app-internal-scaffolds} examines whether changing the inference
harness improves verdict accuracy and verbalized-confidence discrimination
while keeping model parameters fixed.

\begin{table*}[t]
    \centering
    \small
    \setlength{\tabcolsep}{6pt}
    \begin{tabular}{llrr}
        \toprule
        Model & Harness & Acc. (\%) & Verbalized AUROC \\
        \midrule
        \multirow{3}{*}{Qwen3-8B}
        & 0-judge
        & 63.7 & 0.665 \\
        & 3-subjudge
        & 65.3 & 0.634 \\
        & 3-subjudge + RAG
        & \textbf{66.0} & 0.658 \\
        \midrule
        \multirow{3}{*}{Qwen3-14B}
        & 0-judge
        & 65.6 & 0.636 \\
        & 3-subjudge
        & 68.9 & 0.628 \\
        & 3-subjudge + RAG
        & \textbf{70.4} & 0.603 \\
        \bottomrule
    \end{tabular}
    \caption{
    \textbf{Training-free harness comparison on the proprietary task.}
    Model parameters are fixed within each scale, and all harnesses may
    output \textsc{Unsure}.
    Accuracy is computed over all $2{,}634$ examples, counting
    \textsc{Unsure} as incorrect.
    Verbalized AUROC is computed only on non-\textsc{Unsure} judgments;
    its evaluation subset may therefore differ across harnesses.
    Bold marks the highest accuracy within each model scale.
    }
    \label{tab:app-internal-scaffolds}
\end{table*}

The panel-based harnesses reduce abstention and increase all-sample accuracy
at both scales. The highest observed accuracy rises from $63.7\%$ to
$66.0\%$ at 8B and from $65.6\%$ to $70.4\%$ at 14B.
The committed-only AUROC values vary across harnesses, but they cannot
establish a full-population confidence-ranking advantage because different
examples are excluded by abstention.
The all-sample AUROC comparison is therefore needed before concluding
whether changing the harness improves verbalized reliability on this task.

\paragraph{Computational context.}
Table~\ref{tab:app-scaffold-cost} reports the available latency measurements.
Within the harness implementation, the multi-call configurations take longer
than a single direct judge call at both scales.
The post-trained-judge timings are included for context, but use different
batching and timing boundaries; they are not a matched estimate of speedup
or of the incremental cost of the \vera~readout.

\begin{table}[t]
    \centering
    \small
    \setlength{\tabcolsep}{6pt}
    \begin{tabular}{llr}
        \toprule
        Model & Configuration & Time (s/example) \\
        \midrule
        \multirow{4}{*}{Qwen3-8B}
        & 0-judge & 14.04 \\
        & 3-subjudge & 36.21 \\
        & 3-subjudge + RAG & 37.18 \\
        & Post-trained judge$^\dagger$ & 0.24 \\
        \midrule
        \multirow{4}{*}{Qwen3-14B}
        & 0-judge & 12.18 \\
        & 3-subjudge & 43.64 \\
        & 3-subjudge + RAG & 44.91 \\
        & Post-trained judge$^\dagger$ & 0.39 \\
        \bottomrule
    \end{tabular}
    \caption{
    \textbf{Reported inference times on the proprietary task.}
    Harness rows report mean per-example time under the sequential local
    implementation, excluding model loading, file I/O, retrieval-service time,
    and idle time after a shard finishes.
    $^\dagger$Post-trained-judge times are measured end to end over all
    $2{,}634$ examples on one A100-40GB GPU with batch size four, including
    model loading. The timing conventions differ across the two groups.
    }
    \label{tab:app-scaffold-cost}
\end{table}

\subsection{Public Pairwise Harness Results}
\label{app:external-scaffolds}

On MT-Bench-HJ, the three harnesses use the same evaluation set of $6{,}710$
examples. The RAG configuration supplies the top five labeled JudgeLM
examples to the meta-judge, matching the retrieval placement in the
proprietary experiment.
None of the six runs abstains, so the verbalized AUROCs in
Table~\ref{tab:app-external-scaffolds} already include the full set.

\begin{table}[t]
    \centering
    \small
    \setlength{\tabcolsep}{6pt}
    \begin{tabular}{llrrr}
        \toprule
        Model & Harness & Unsure (\%) & Acc. (\%) & Verbalized AUROC \\
        \midrule
        \multirow{3}{*}{Qwen3-8B}
        & 0-judge & 0.0 & \textbf{62.0} & 0.533 \\
        & 3-subjudge$^\dagger$ & 0.0 & 60.6 & 0.551 \\
        & 3-subjudge + RAG$^\dagger$ & 0.0 & 59.6 & \textbf{0.565} \\
        \midrule
        \multirow{3}{*}{Qwen3-14B}
        & 0-judge & 0.0 & 62.8 & 0.557 \\
        & 3-subjudge$^\dagger$ & 0.0 & 64.2 & \textbf{0.566} \\
        & 3-subjudge + RAG$^\dagger$ & 0.0 & \textbf{64.4} & 0.558 \\
        \bottomrule
    \end{tabular}
    \caption{
    \textbf{Training-free harness comparison on MT-Bench-HJ.}
    All metrics use all $6{,}710$ examples; ties are valid verdicts and are
    retained. Verbalized AUROC uses the explicit high, medium, low, or unsure
    confidence label. For otherwise valid truncated outputs, confidence is
    recovered only from its named field.
    The 0-judge rows use temperature zero; $^\dagger$stochastic panel rows
    are single-run results.
    Bold marks the highest accuracy or AUROC within each model scale.
    }
    \label{tab:app-external-scaffolds}
\end{table}

Changing the harness affects stated-confidence ranking, but its effect is
not aligned uniformly with accuracy.
At 8B, the RAG panel raises verbalized AUROC from $0.533$ to $0.565$ while
reducing accuracy from $62.0\%$ to $59.6\%$.
At 14B, the RAG panel improves accuracy from $62.8\%$ to $64.4\%$, while
verbalized AUROC changes only from $0.557$ to $0.558$; the panel without
RAG yields the highest AUROC, $0.566$.
Across these configurations, verbalized AUROC remains between $0.533$ and
$0.566$. Additional inference can therefore change the reported confidence
signal, but does not consistently produce strong correctness discrimination
in this evaluation.
This motivates examining internal reliability information alongside
output-level confidence, rather than assuming that a more elaborate harness
resolves the reliability-estimation problem.

For task-performance context, the \vera-adapted judges in
Table~\ref{tab:main-cross-dataset} achieve MT-Bench-HJ accuracies of
$66.69\%$ and $67.73\%$, exceeding the strongest harness accuracy at each
scale by relative margins of $7.56\%$ and $5.17\%$, respectively.
This compares complete trained judges with fixed-parameter harnesses; it does
not isolate the effect of the reliability readout from the effect of training.

\section{Cross-Format Generalization and Benchmark Placement}
\label{app:mixed-format}

We extend the public comparison in Table~\ref{tab:main-cross-dataset} with
benchmark-specific results and selected published references.
Our results in the tables in this section report means and sample standard
deviations over three seeds, consistent with
Table~\ref{tab:main-cross-dataset}.
The primary comparison is between IFT, DPO, and \vera-guided periodic
adaptation at the same model scale.
Published references provide context rather than a parameter- or
training-compute-matched comparison.
The reference set follows evaluator categories, spanning specialized and
general-purpose open- and closed-source LLM judges while excluding dedicated
reward models and references classified as preference-optimization-based
evaluators. The Safety-repaired checkpoints in
Table~\ref{tab:app-rb2-safety-repair} are not included in these tables.

\subsection{Pairwise Preference Judging: Auto-J}
\label{app:autoj-results}

Auto-J Eval-P~\citep{li2024generativejudge} evaluates three-way preference
judgments with genuine ties.
Agreement requires predictions under both response orders to be consistent
after label inversion and to match the human reference; consistency measures
agreement between the two orders regardless of correctness.

\begin{table}[t]
    \centering
    \small
    \setlength{\tabcolsep}{6pt}
    \begin{tabular}{lrr}
        \toprule
        Evaluator & Agreement (\%) & Consistency (\%) \\
        \midrule
        \multicolumn{3}{l}{\emph{Published Auto-J results}} \\
        LLaMA-2-13B-Chat & 29.81 & 48.56 \\
        WizardLM-13B-v1.2 & 36.35 & 57.69 \\
        Vicuna-13B-v1.5 & 39.22 & 62.07 \\
        PandaLM & 39.44 & 66.88 \\
        Claude-2 & 42.60 & 63.43 \\
        ChatGPT & 42.74 & 62.43 \\
        LLaMA-2-70B-Chat & 46.12 & 69.90 \\
        Auto-J-13B & 54.96 & 83.41 \\
        GPT-4 & 62.28 & 86.28 \\
        \midrule
        \multicolumn{3}{l}{\emph{Qwen3-8B}} \\
        IFT & $58.93\pm0.61$ & $\mathbf{89.94\pm1.25}$ \\
        DPO & $58.93\pm0.49$ & $89.45\pm0.97$ \\
        \vera & $\mathbf{59.85\pm1.01}$ & $85.84\pm2.29$ \\
        Relative gain over best baseline & $+1.56$ & $-4.56$ \\
        \midrule
        \multicolumn{3}{l}{\emph{Qwen3-14B}} \\
        IFT & $60.69\pm0.17$ & $\mathbf{90.26\pm0.23}$ \\
        DPO & $60.85\pm0.25$ & $89.95\pm0.19$ \\
        \vera & $\mathbf{62.54\pm0.43}$ & $87.42\pm0.45$ \\
        Relative gain over best baseline & $+2.78$ & $-3.15$ \\
        \bottomrule
    \end{tabular}
    \caption{
    \textbf{Auto-J Eval-P results.}
    Published references are taken from the Auto-J comparison.
    IFT, DPO, and \vera~denote JudgeLM instruction fine-tuning, subsequent
    Arena DPO, and \vera-guided periodic adaptation, respectively.
    Entries report means and sample standard deviations over three seeds.
    Gains are relative improvements (\%) computed from displayed means, comparing
    \vera~with the stronger of IFT and DPO for each metric and model scale.
    Bold marks the best mean of these three methods.
    }
    \label{tab:app-autoj-leaderboard}
\end{table}

\vera~improves mean agreement over the strongest post-training baseline by
$1.56\%$ at 8B and $2.78\%$ at 14B in relative terms.
Mean consistency is lower by $4.56\%$ and $3.15\%$ relative to the strongest baseline,
respectively. Thus, adaptation improves agreement with reference judgments
but does not improve every aspect of response-order robustness.

\subsection{Multi-Way Selection: RewardBench 2}
\label{app:rb2-results}

RewardBench~2~\citep{malik2026rewardbench2} tests response selection across
six subsets. Our generative judges select among four responses and are scored
by averaging item accuracy over four cyclic rotations.
Table~\ref{tab:app-rb2-leaderboard} reports the per-subset results, allowing
improvements and retention losses to be examined separately.

\begin{table*}[t]
    \centering
    \small
    \setlength{\tabcolsep}{5pt}
    \resizebox{\textwidth}{!}{%
    \begin{tabular}{lrrrrrrr}
        \toprule
        Evaluator & Avg.$^\dagger$ & Fact. & Prec. IF
        & Math & Safety & Focus & Ties$^\dagger$ \\
        \midrule
        \multicolumn{8}{l}{\emph{Published RewardBench 2 results}} \\
        LMUnit-Qwen2.5-72B
        & 82.1 & 87.2 & 54.4 & 72.7 & 91.3 & 96.8 & 90.1 \\
        LMUnit-Llama3.1-70B
        & 80.5 & 84.6 & 48.8 & 71.6 & 90.7 & 97.0 & 90.6 \\
        Gemini-2.5-Pro
        & 79.5 & 75.5 & 61.9 & 89.8 & 88.1 & 80.5 & 81.1 \\
        Gemini-2.5-Flash
        & 77.7 & 67.4 & 57.5 & 85.2 & 90.9 & 84.1 & 80.9 \\
        Claude Opus 4
        & 76.5 & 82.7 & 41.9 & 74.9 & 89.5 & 86.2 & 83.7 \\
        Claude 3.7 Sonnet
        & 75.4 & 73.3 & 54.4 & 75.0 & 90.3 & 92.1 & 67.2 \\
        GPT-4.1
        & 72.3 & 82.9 & 39.7 & 65.2 & 87.3 & 73.4 & 85.4 \\
        GPT-4o
        & 64.9 & 56.8 & 33.1 & 62.3 & 86.2 & 72.9 & 78.2 \\
        \midrule
        \multicolumn{8}{l}{\emph{Qwen3-8B}} \\
        IFT
        & $\mathbf{66.87\pm0.04}$ & $51.97\pm0.49$ & $30.84\pm0.55$ & $66.54\pm0.15$ & $\mathbf{81.43\pm0.82}$ & $76.91\pm0.18$ & $93.45\pm0.39$ \\
        DPO
        & $66.36\pm0.25$ & $51.47\pm0.20$ & $31.01\pm0.35$ & $66.68\pm0.27$ & $76.06\pm1.44$ & $79.09\pm0.51$ & $93.88\pm0.25$ \\
        \vera
        & $65.22\pm0.07$ & $\mathbf{55.96\pm0.59}$ & $\mathbf{35.00\pm0.62}$ & $\mathbf{70.04\pm0.75}$ & $50.36\pm0.88$ & $\mathbf{84.50\pm0.93}$ & $\mathbf{95.50\pm0.36}$ \\
        \midrule
        \multicolumn{8}{l}{\emph{Qwen3-14B}} \\
        IFT
        & $67.06\pm0.14$ & $50.49\pm0.40$ & $30.26\pm1.02$ & $67.70\pm0.82$ & $\mathbf{81.09\pm0.37}$ & $79.88\pm0.83$ & $93.04\pm0.13$ \\
        DPO
        & $66.31\pm0.17$ & $49.65\pm0.45$ & $30.28\pm0.32$ & $67.58\pm0.64$ & $75.27\pm0.83$ & $81.39\pm0.51$ & $93.70\pm0.50$ \\
        \vera
        & $\mathbf{67.35\pm0.28}$ & $\mathbf{56.39\pm0.32}$ & $\mathbf{34.84\pm0.94}$ & $\mathbf{70.80\pm1.08}$ & $57.29\pm1.60$ & $\mathbf{89.82\pm0.83}$ & $\mathbf{94.85\pm0.88}$ \\
        \bottomrule
    \end{tabular}}
    \caption{
    \textbf{RewardBench 2 results (\%).}
    Our judges use position-averaged discrete best-of-four accuracy over four
    cyclic rotations. Entries report means and sample standard deviations over three seeds.
    Bold marks the best mean of IFT, DPO, and \vera~within
    each model scale.
    $^\dagger$The published Ties score incorporates score margins, whereas
    our Ties evaluation uses discrete selection. Consequently, neither the
    Ties column nor an average incorporating it is directly comparable between
    published-reference and Qwen3 blocks.
    }
    \label{tab:app-rb2-leaderboard}
\end{table*}

Within the Qwen3 comparisons, \vera~improves all five non-Safety subset
scores over both IFT and DPO at each scale.

{\color{reviewretention}
These transfer gains coexist with domain-specific retention losses.
The public training pipeline uses JudgeLM judgments and Arena preferences,
without a dedicated Safety-training objective; RewardBench~2 Safety
therefore probes transfer beyond the adaptation target.
The IFT scores document Safety-judging capability already present before
periodic adaptation.
On RewardBench~2, mean Safety accuracy falls from $81.43\%$ under IFT to
$50.36\%$ after adaptation at 8B, and from $81.09\%$ to $57.29\%$ at 14B
(Table~\ref{tab:app-rb2-leaderboard}).
The macro-average decreases from $66.87\%$ to $65.22\%$ at 8B and
increases from $67.06\%$ to $67.35\%$ at 14B.
Auto-J response-order consistency also falls from $89.94\%$ to $85.84\%$
at 8B and from $90.26\%$ to $87.42\%$ at 14B
(Table~\ref{tab:app-autoj-leaderboard}).
The separate Safety-repair experiment uses additional labeled target
judgments and produces new checkpoints
(Appendix~\ref{app:rb2-safety-retrieval}); it measures targeted recovery,
not retention by the original adaptation procedure.
}

\subsection{Objective Correctness: JudgeBench}
\label{app:judgebench-results}

JudgeBench~\citep{tan2025judgebench} evaluates pairwise judgments of
objective correctness without ties.
Table~\ref{tab:app-judgebench-leaderboard} uses its $350$-item GPT-4o
subset and two-order aggregation.

\begin{table}[t]
    \centering
    \small
    \setlength{\tabcolsep}{6pt}
    \begin{tabular}{lr}
        \toprule
        Evaluator & Accuracy (\%) \\
        \midrule
        \multicolumn{2}{l}{\emph{Published JudgeBench results}} \\
        o3-mini (high), Arena-Hard & 80.86 \\
        Claude-3.5-Sonnet, Arena-Hard & 64.29 \\
        GPT-4o, Arena-Hard & 56.57 \\
        Skywork-LLaMA-3.1-8B judge & 53.43 \\
        VertexAI Gemini-1.5-Pro & 44.57 \\
        Prometheus2-8x7B & 40.29 \\
        Auto-J & 36.57 \\
        JudgeLM-33B & 35.71 \\
        PandaLM & 13.14 \\
        \midrule
        \multicolumn{2}{l}{\emph{Qwen3-8B}} \\
        IFT & $56.48\pm0.59$ \\
        DPO & $57.06\pm0.32$ \\
        \vera & $\mathbf{57.22\pm1.86}$ \\
        Relative gain over best baseline & $+0.28$ \\
        \midrule
        \multicolumn{2}{l}{\emph{Qwen3-14B}} \\
        IFT & $60.20\pm1.17$ \\
        DPO & $60.46\pm1.31$ \\
        \vera & $\mathbf{62.08\pm1.02}$ \\
        Relative gain over best baseline & $+2.68$ \\
        \bottomrule
    \end{tabular}
    \caption{
    \textbf{JudgeBench objective-correctness results.}
    Scores use the $350$-item GPT-4o subset with two-order aggregation.
    Published references provide contextual comparisons with prompted and
    fine-tuned LLM judges.
    Entries report means and sample standard deviations over three seeds.
    Gains are relative improvements (\%) over the stronger of IFT and DPO,
    calculated from the displayed two-decimal means.
    Bold marks the best mean of our three methods within each model scale.
    }
    \label{tab:app-judgebench-leaderboard}
\end{table}

\vera~reaches mean accuracy of $57.22\%$ at 8B and $62.08\%$ at 14B,
exceeding the stronger post-training baseline by relative margins of $0.28\%$ and $2.68\%$, respectively.
The 8B improvement is small; the 14B result provides a larger numerical gain
on a task whose labels do not include ties.
These values are reported as observed differences rather than evidence of
statistical significance.

\subsection{Adversarial Instruction Following: LLMBar}
\label{app:llmbar-results}

LLMBar~\citep{zeng2024llmbar} evaluates whether a judge identifies the
response that follows the instruction despite misleading surface cues.
We report the macro-average over its Neighbor, GPTInst, GPTOut, and Manual
subsets, averaging predictions over both response orders.

\begin{table}[t]
    \centering
    \small
    \setlength{\tabcolsep}{6pt}
    \begin{tabular}{lr}
        \toprule
        Evaluator & Adversarial Acc. (\%) \\
        \midrule
        \multicolumn{2}{l}{\emph{Published LLMBar results}} \\
        ChatGPT, vanilla & 32.0 \\
        ChatGPT, enhanced prompt & 38.9 \\
        LLaMA-2-70B, vanilla & 32.6 \\
        LLaMA-2-70B, enhanced prompt & 43.4 \\
        PaLM2, vanilla & 60.7 \\
        PaLM2, Rules+Metrics+Ref & 69.5 \\
        GPT-4, vanilla & 73.1 \\
        GPT-4, Rules+Metrics+Ref & 82.8 \\
        \midrule
        \multicolumn{2}{l}{\emph{Qwen3-8B}} \\
        IFT & $56.06\pm0.37$ \\
        DPO & $59.63\pm0.12$ \\
        \vera & $\mathbf{64.95\pm2.40}$ \\
        Relative gain over best baseline & $+8.92$ \\
        \midrule
        \multicolumn{2}{l}{\emph{Qwen3-14B}} \\
        IFT & $54.95\pm0.58$ \\
        DPO & $57.05\pm1.25$ \\
        \vera & $\mathbf{70.18\pm1.12}$ \\
        Relative gain over best baseline & $+23.01$ \\
        \bottomrule
    \end{tabular}
    \caption{
    \textbf{LLMBar adversarial instruction-following results.}
    Accuracy is macro-averaged across four adversarial subsets using
    position-averaged evaluation over both response orders.
    Published references retain multiple prompting configurations because
    LLMBar explicitly compares these protocols.
    Entries report means and sample standard deviations over three seeds.
    Gains are relative improvements (\%) over the stronger of IFT and DPO;
    bold marks the best mean of our three methods within each model scale.
    }
    \label{tab:app-llmbar-leaderboard}
\end{table}

Across three seeds, \vera~improves mean adversarial accuracy from
$59.63\%$ to $64.95\%$ at 8B and from $57.05\%$ to $70.18\%$ at 14B,
relative gains of $8.92\%$ and $23.01\%$ over the stronger baseline.
These are the largest gains among the four public benchmarks in
Table~\ref{tab:main-cross-dataset}.
Together with the Auto-J and JudgeBench results, they show that the benefits
of periodic adaptation extend beyond the preference data used for updating
the judge. The Safety and consistency results also show why this transfer
should not be interpreted as uniform improvement across all capabilities.

\end{document}